\documentclass{article}

\PassOptionsToPackage{numbers, compress}{natbib}

\usepackage[main, final]{neurips_2026}

\usepackage{amsmath,amsfonts,bm}

\def\eqref#1{equation~\ref{#1}}

\def\1{\bm{1}}

\def\rmA{{\mathbf{A}}}
\def\rmB{{\mathbf{B}}}

\def\rmD{{\mathbf{D}}}
\def\rmE{{\mathbf{E}}}

\def\rmI{{\mathbf{I}}}

\def\rmK{{\mathbf{K}}}

\def\rmM{{\mathbf{M}}}

\def\rmO{{\mathbf{O}}}

\def\rmQ{{\mathbf{Q}}}
\def\rmR{{\mathbf{R}}}
\def\rmS{{\mathbf{S}}}
\def\rmT{{\mathbf{T}}}
\def\rmU{{\mathbf{U}}}
\def\rmV{{\mathbf{V}}}
\def\rmW{{\mathbf{W}}}

\def\rmY{{\mathbf{Y}}}
\def\rmZ{{\mathbf{Z}}}

\def\vone{{\bm{1}}}
\def\valpha{{\bm{\alpha}}}

\def\va{{\bm{a}}}
\def\vb{{\bm{b}}}

\def\ve{{\bm{e}}}

\def\vg{{\bm{g}}}

\def\vk{{\bm{k}}}

\def\vo{{\bm{o}}}

\def\vq{{\bm{q}}}
\def\vr{{\bm{r}}}

\def\vu{{\bm{u}}}
\def\vv{{\bm{v}}}
\def\vw{{\bm{w}}}
\def\vx{{\bm{x}}}
\def\vy{{\bm{y}}}
\def\vz{{\bm{z}}}

\DeclareMathAlphabet{\mathsfit}{\encodingdefault}{\sfdefault}{m}{sl}
\SetMathAlphabet{\mathsfit}{bold}{\encodingdefault}{\sfdefault}{bx}{n}

\usepackage[utf8]{inputenc} 
\usepackage[T1]{fontenc}    
\usepackage{hyperref}       
\usepackage{url}            
\usepackage{booktabs}       
\usepackage{amsfonts}       
\usepackage{nicefrac}       
\usepackage{microtype}      
\usepackage{xcolor}         
\PassOptionsToPackage{numbers, compress}{natbib}
\usepackage{amsmath,amssymb}
\usepackage{tikz}
\usetikzlibrary{positioning,arrows.meta,fit,backgrounds,calc}

\providecommand{\rmA}{\mathbf{A}}
\providecommand{\rmB}{\mathbf{B}}
\providecommand{\rmD}{\mathbf{D}}
\providecommand{\rmE}{\mathbf{E}}
\providecommand{\rmI}{\mathbf{I}}
\providecommand{\rmK}{\mathbf{K}}
\providecommand{\rmM}{\mathbf{M}}
\providecommand{\rmO}{\mathbf{O}}
\providecommand{\rmQ}{\mathbf{Q}}
\providecommand{\rmR}{\mathbf{R}}
\providecommand{\rmS}{\mathbf{S}}
\providecommand{\rmT}{\mathbf{T}}
\providecommand{\rmU}{\mathbf{U}}
\providecommand{\rmV}{\mathbf{V}}
\providecommand{\rmW}{\mathbf{W}}
\providecommand{\rmY}{\mathbf{Y}}
\providecommand{\rmZ}{\mathbf{Z}}
\providecommand{\vk}{\bm{k}}
\providecommand{\vv}{\bm{v}}
\providecommand{\vq}{\bm{q}}
\providecommand{\vo}{\bm{o}}
\providecommand{\vw}{\bm{w}}
\providecommand{\vu}{\bm{u}}

\usepackage{hyperref}
\usepackage{url}

\usepackage{microtype}
\usepackage{amsthm}

\usepackage{subfigure}
\usepackage{booktabs} 
\usepackage{breakurl}
\hypersetup{
           breaklinks=true,   
           colorlinks=true,   
           pdfusetitle=true,  
        }

\usepackage{tikz}
\usepackage{pifont}
\usepackage{pgfplots}
\pgfplotsset{compat=1.18}
\usepackage{amsthm}
\usepackage{multirow}
\usepackage{scalefnt}
\usepackage{minted}
\usepackage{listings}
\usepackage{float}
\usepackage[utf8]{inputenc} 
\usepackage[T1]{fontenc}    
\usepackage{hyperref}       
\usepackage{url}            
\usepackage{booktabs}       
\usepackage{amsfonts}       
\usepackage{nicefrac}       
\usepackage{microtype}      
\usepackage[table]{xcolor}         
\usepackage{tikz}
\usepackage{caption}
\usepackage{subcaption}
\usepackage{inconsolata}
\usepackage{graphicx}
\usepackage{amssymb}
\usepackage{bm}
\usepackage{bbm}
\usepackage{amsmath}  %
\usepackage{amsthm}
\usepackage{xcolor, color}
\usepackage{textcomp}
\usepackage{multicol}
\usepackage{multirow}
\usepackage{mathtools}
\usepackage{wrapfig}
\usepackage{mathtools}
\usepackage{tcolorbox}
\usepackage{pgfplots}

\usepgflibrary{patterns}
\usetikzlibrary{arrows.meta,positioning,calc,shapes.geometric,decorations.text,patterns,patterns.meta,fit,backgrounds,chains,shadows,math,circuits.ee.IEC,decorations.pathmorphing, decorations.pathreplacing, decorations.shapes}
\usepgfplotslibrary{fillbetween}

\usepackage{amsmath}
\usepackage{amssymb}
\usepackage{mathtools}
\usepackage{amsthm}
\usepackage{fontawesome}
\usepackage{hyperref}
\definecolor{darkblue}{rgb}{0, 0, 0.5}
\hypersetup{colorlinks, citecolor=darkblue, linkcolor=darkblue, urlcolor=darkblue}

\usepackage[utf8]{inputenc} 
\usepackage[T1]{fontenc}    
\usepackage{url}            
\usepackage{booktabs}       
\usepackage{amsfonts}       
\usepackage{nicefrac}       
\usepackage{microtype}      

\usepackage{wrapfig,lipsum}

\usepackage{xcolor}         
\usepackage{amsmath}         
\usepackage[labelfont=bf,font=small]{caption}

\usetikzlibrary{arrows.meta,positioning,calc,shapes.geometric,decorations.text,patterns,patterns.meta,fit,backgrounds,chains,shadows,math,circuits.ee.IEC,decorations.pathmorphing, decorations.pathreplacing, decorations.shapes}

\definecolor{brickred}{HTML}{b92622}
\definecolor{midnightblue}{HTML}{005c7f}
\definecolor{salmon}{HTML}{f1958d}
\definecolor{burntorange}{HTML}{f19249}
\definecolor{junglegreen}{HTML}{4dae9d}
\definecolor{forestgreen}{HTML}{499c5e}
\definecolor{pinegreen}{HTML}{3d8a75}
\definecolor{seagreen}{HTML}{6bc1a2}
\definecolor{limegreen}{HTML}{97c65a}
\definecolor{violet}{HTML}{8f00ff}
\definecolor{pastelviolet}{HTML}{cb99c9}
\definecolor{darkcyan}{HTML}{008B8B}

\def\rmU{{\mathbf{U}}}

\def\vw{\mathbf{w}}
\def\vu{\mathbf{u}}

\definecolor{bred}{RGB}{250, 82, 82}
\definecolor{borange}{RGB}{253, 126, 20}
\definecolor{byellow}{RGB}{250, 176, 5}
\definecolor{bgreen}{RGB}{116, 184, 22}
\definecolor{bblue}{RGB}{250, 176, 5}
\definecolor{bindigo}{RGB}{76, 110, 245}
\definecolor{bcyan}{RGB}{59, 201, 219}
\definecolor{bteal}{RGB}{99, 230, 190}

\usepackage{graphicx}

\title{Gated DeltaNet-2: Decoupling Erase and Write in Linear Attention}

\author{%
  Ali Hatamizadeh \quad Yejin Choi \quad Jan Kautz \\
  NVIDIA \\
  \texttt{\{ahatamizadeh,yejinc,jkautz\}@nvidia.com} \\
}

\begin{document}

\maketitle

\begin{abstract}
Linear attention replaces the unbounded cache of softmax attention with a fixed-size recurrent state, reducing sequence mixing to linear time and decoding to constant memory. The hard part is not just what to forget, but how to edit this compressed memory without scrambling existing associations. Delta-rule models subtract the current read before writing a new value, and Kimi Delta Attention (KDA) sharpens forgetting with channel-wise decay. But the active edit still uses a single scalar gate to control two different things, how much old content to erase on the key side and how much new content to commit on the value side. We introduce Gated DeltaNet-2, which generalizes both Gated DeltaNet and KDA by inheriting adaptive forgetting and channel-wise decay while addressing their shared limitation, the scalar tie between erasing and writing. Gated Delta Rule-2 separates these roles with a channel-wise erase gate $\vb_t$ and a channel-wise write gate $\vw_t$, reducing to KDA when both gates collapse to the same scalar and to Gated DeltaNet when the decay also collapses. We derive a fast-weight update view, a chunkwise WY algorithm with channel-wise decay absorbed into asymmetric erase factors, and a gate-aware backward pass that preserves efficient parallel training. At 1.3B parameters trained on 100B FineWeb-Edu tokens, Gated DeltaNet-2 achieves the strongest overall results among Mamba-2, Gated DeltaNet, KDA, and Mamba-3 variants across language modeling, commonsense reasoning, and retrieval. Its advantage is most pronounced on long-context RULER needle-in-a-haystack benchmarks, where it improves the evaluated multi-key retrieval setting and remains strong in both recurrent and hybrid settings.\\
Code: \url{https://github.com/NVlabs/GatedDeltaNet-2}
\end{abstract}

\section{Introduction}
\label{sec_introduction}

The Transformer architecture has become the dominant backbone for large language models because self-attention gives each token direct access to its history and maps naturally to parallel training on modern accelerators. Its cost, however, still grows quadratically with sequence length. This cost becomes a central obstacle for long-context training and high-throughput inference, where the model must repeatedly process histories that are much longer than the dimension of a single attention head.

Linear recurrent attention takes a different path. It replaces the explicit attention matrix with a fixed-size recurrent state, turning sequence mixing into a linear-time recurrence whose memory does not grow with context length~\citep{katharopoulos_transformers_2020}. The appeal is clear, but so is the constraint. The state is a compressed key-value memory, and long contexts force many associations to share the same finite space, making exact retrieval difficult~\citep{linear-xmr-fastweight,zoology,arora_simple_2024,jelassi_repeat_2024,wen_rnns_2024,akyurek_-context_2024}. Recent work has improved this memory by giving the recurrence more control over what persists. Mamba-2 uses data-dependent decay to regulate the memory horizon~\citep{pmlr-v235-dao24a}. DeltaNet replaces additive writes with the delta rule, enabling targeted overwrite of the association addressed by the current key~\citep{widrow_adaptive_1988,linear-xmr-fastweight,yang2024parallelizing}. Gated DeltaNet combines the delta rule with a learned decay gate, giving the state both global forgetting and targeted editing~\citep{yang2025gated}. Kimi Delta Attention (KDA) refines the decay side with channel-wise forgetting over the key dimension~\citep{team2025kimi}. In parallel, Mamba-3 advances the state-space route through exponential-trapezoidal discretization, complex-valued state transitions, and a multi-input, multi-output formulation for stronger and more efficient recurrence~\citep{lahoti2026mamba3}. These advances have pushed recurrent linear models forward, while making the remaining bottleneck in delta-rule memory more visible. The active edit still uses one scalar gate to control both erasing old content and writing new content.

We propose \emph{Gated DeltaNet-2}, a recurrent attention layer that decouples erase and write in the delta rule. The scalar tie is a modeling restriction because erasing and writing act on different axes of the state. Erasing is a key-side operation that decides which coordinates of the old read should be removed, while writing is a value-side operation that decides which coordinates of the incoming value should be committed. Gated DeltaNet-2 preserves KDA's channel-wise decay, but replaces the tied scalar delta gate with a channel-wise erase gate on the key axis and a channel-wise write gate on the value axis. The model can clear broad context through decay, remove selected stale associations through erase, and insert only the value channels that should persist through write. When the erase and write gates are tied to the same scalar, Gated DeltaNet-2 recovers KDA. If the decay is tied to a scalar as well, it recovers Gated DeltaNet.

This change preserves the efficient training path. By absorbing cumulative channel-wise decay into the rank-one erase factors, the recurrence admits a compact WY form with the same high-level chunkwise structure used by efficient delta-rule kernels~\citep{bischof_wy_1985,hua_transformer_2022,sun2023retentive,yang_gated_2023}. The main text gives the modeling equations and the chunkwise algorithm. Kernel-level details are deferred to the supplement.

Empirically, Gated DeltaNet-2 improves the recurrent attention frontier, with the clearest gains on long-context retrieval. On the RULER needle-in-a-haystack tasks in Table~\ref{tab_niah_results}, it remains strong as context length grows and is especially effective on the evaluated multi-key case where a fixed-size state must separate competing associations. This advantage also appears in real-world recall, where Gated DeltaNet-2 gives the strongest overall retrieval profile in both recurrent and hybrid settings. Together with gains in language modeling, commonsense reasoning and in-context retrieval, these results suggest that decoupling the active memory edit directly targets the main pressure point of fixed-state recurrence, interference among many compressed associations.

\section{Preliminary}
\label{sec_preliminary}

\subsection{Linear attention as a recurrent state}

We work with one attention head and omit layer indices. Let $\vq_t, \vk_t \in \mathbb{R}^{d_k}$ and $\vv_t \in \mathbb{R}^{d_v}$ denote the query, key, and value at position $t$. A recurrent linear attention layer stores a matrix state $\rmS_t \in \mathbb{R}^{d_k \times d_v}$ and reads it with the query,
\begin{align}
    \rmS_t &= \rmS_{t-1} + \vk_t \vv_t^\top,
    &
    \vo_t &= \rmS_t^\top \vq_t .
\label{eq_linear_attention_recurrence}
\end{align}
This is the recurrent form of linear attention~\citep{katharopoulos_transformers_2020}. Expanding the recurrence over a length $L$ sequence gives the familiar causal matrix form
\begin{align}
    \rmO = (\rmQ\rmK^\top \odot \rmM)\rmV,
\label{eq_linear_attention_parallel}
\end{align}
where $\rmM$ is the causal mask. The state has fixed size in $L$, and the parallel form replaces tokenwise recurrence with matrix multiplication. The limitation is equally direct. Every outer product is added to the state and none is removed, so old associations remain until they are overwritten indirectly by later superposition.

\paragraph{Chunkwise form}
Efficient linear recurrent layers use a chunkwise schedule during training~\citep{hua_transformer_2022, sun2023retentive, yang_gated_2023}. Split the sequence into chunks of size $C$. For chunk $n$, let $\rmQ_{[n]}, \rmK_{[n]}, \rmV_{[n]}$ be the query, key, and value blocks, and let $\rmS_{[n]}$ be the state at the start of the chunk. Partial expansion gives
\begin{align}
    \rmS_{[n+1]} &= \rmS_{[n]} + \rmK_{[n]}^\top \rmV_{[n]},
    &
    \rmO_{[n]} &= \rmQ_{[n]}\rmS_{[n]} + (\rmQ_{[n]}\rmK_{[n]}^\top \odot \rmM_C)\rmV_{[n]} .
\label{eq_linear_attention_chunk}
\end{align}
The recurrence remains only across chunks, while all token interactions inside a chunk are expressed as dense matrix products. With a fixed $C$, this keeps linear complexity in sequence length and maps well to tensor cores.

\subsection{Forgetting and overwriting}

Mamba-2 adds a data-dependent scalar decay before each write~\citep{pmlr-v235-dao24a},
\begin{align}
    \rmS_t = \alpha_t\rmS_{t-1} + \vk_t\vv_t^\top,
    \qquad
    \alpha_t \in (0, 1] .
\label{eq_mamba2_recurrence}
\end{align}
The decay gives the model a global forgetting operation. If $\gamma_t = \prod_{i=1}^t \alpha_i$, then each earlier write is read at time $t$ with factor $\gamma_t / \gamma_i$. This yields a decay-aware attention mask and preserves the chunkwise structure of Eq.~\ref{eq_linear_attention_chunk}.

DeltaNet instead gives the state an active edit operation~\citep{widrow_adaptive_1988, linear-xmr-fastweight, yang2024parallelizing}. Before writing $\vv_t$, the model reads the value currently associated with $\vk_t$ and subtracts it from the state. With a scalar step size $\beta_t \in [0,1]$, the update is
\begin{align}
    \rmS_t
    &= \rmS_{t-1} + \beta_t\vk_t(\vv_t - \rmS_{t-1}^\top\vk_t)^\top
     = (\rmI - \beta_t\vk_t\vk_t^\top)\rmS_{t-1} + \beta_t\vk_t\vv_t^\top .
\label{eq_delta_recurrence}
\end{align}
When $\|\vk_t\|_2=1$, the matrix $\vk_t\vk_t^\top$ is a projector, so $\beta_t=1$ overwrites the association at key $\vk_t$ and $\beta_t=0$ leaves it unchanged. In the fast-weight view~\citep{Irie2022TheDF, ttt}, Eq.~\ref{eq_delta_recurrence} is one online gradient step on the local regression loss $\frac{1}{2}\|\rmS^\top\vk_t - \vv_t\|_2^2$.

Gated DeltaNet combines these two operations~\citep{yang2025gated},
\begin{align}
    \rmS_t
    = \alpha_t(\rmI - \beta_t\vk_t\vk_t^\top)\rmS_{t-1} + \beta_t\vk_t\vv_t^\top .
\label{eq_gdn_recurrence}
\end{align}
The decay clears the state uniformly, while the delta rule edits a selected association. This is a useful division of labor, but both gates are scalar per head.

KDA refines the decay side by replacing the scalar $\alpha_t$ with a channel-wise vector $\valpha_t \in (0,1]^{d_k}$~\citep{team2025kimi}. With $\rmD_t = \operatorname{Diag}(\valpha_t)$, its update can be written as
\begin{align}
    \rmS_t
    = (\rmI - \beta_t\vk_t\vk_t^\top)\rmD_t\rmS_{t-1} + \beta_t\vk_t\vv_t^\top .
\label{eq_kda_recurrence}
\end{align}
KDA lets each key channel decay at its own rate and retains the efficient WY-based chunkwise algorithm of DeltaNet~\citep{bischof_wy_1985, yang2024parallelizing}. Yet the active gate $\beta_t$ is still a single scalar. It controls both how much old content is erased from the read direction and how much new value is written. Gated DeltaNet-2 starts from this remaining tie.

\section{Gated DeltaNet-2}
\label{sec_method}

\subsection{Decoupling erase and write}
\label{subsec_decoupling}

KDA refines Gated DeltaNet by making the decay channel-wise, but the scalar $\beta_t$ in Eq.~\ref{eq_kda_recurrence} still carries two decisions that need not agree. One decision lives on the key side and determines which coordinates of the current read should be erased. The other lives on the value side and determines which coordinates of the candidate value should be written. Treating both decisions as one scalar is a restriction of the update, not a requirement of the delta rule.

Gated DeltaNet-2 separates the two decisions through \emph{Gated Delta Rule-2}. Let
\begin{align}
    \ve_t &= \vb_t \odot \vk_t,
    &
    \vz_t &= \vw_t \odot \vv_t,
\label{eq_gdn2_e_z}
\end{align}
where $\vb_t \in [0,1]^{d_k}$ is the erase gate and $\vw_t \in [0,1]^{d_v}$ is the write gate. The erase gate weights the key coordinates used to read old content, while the write gate weights the value coordinates being inserted. Let $\rmD_t=\operatorname{Diag}(\valpha_t)$. Applying decay before the active edit gives
\begin{align}
    \bar{\rmS}_t &= \rmD_t\rmS_{t-1},
    &
    \vr_t &= \bar{\rmS}_t^\top\ve_t,
    &
    \rmS_t &= \bar{\rmS}_t + \vk_t(\vz_t - \vr_t)^\top .
\label{eq_gdn2_residual}
\end{align}
Equivalently,
\begin{equation}
    \boxed{
    \rmS_t
    = \bigl(\rmI - \vk_t(\vb_t \odot \vk_t)^\top\bigr)\rmD_t\rmS_{t-1}
    + \vk_t(\vw_t \odot \vv_t)^\top
    }
\label{eq_gdn2_recurrence}
\end{equation}
We refer to Eq.~\ref{eq_gdn2_recurrence} as \emph{Gated Delta Rule-2}. The output is $\vo_t = \rmS_t^\top\vq_t$.
The left factor of the erase matrix remains $\vk_t$, which preserves the write direction of the delta rule. The right factor becomes $\vb_t \odot \vk_t$, which makes the read direction channel selective. The write term becomes $\vk_t\vz_t^\top$, which makes the value update channel selective.

Gated Delta Rule-2 recovers KDA exactly when $\vb_t = \beta_t\mathbf{1}_{d_k}$ and $\vw_t = \beta_t\mathbf{1}_{d_v}$. It recovers Gated DeltaNet by further setting $\valpha_t = \alpha_t\mathbf{1}_{d_k}$. Thus the model preserves the known scalar-gated updates as tied subspaces, while learning outside those subspaces when erase and write require different channel structure.

The layer produces the two gates with independent projections of the token representation,
\begin{align}
    \vb_t &= \sigma(\mathbf{W}_b\vx_t),
    &
    \vw_t &= \sigma(\mathbf{W}_w\vx_t) .
\label{eq_gdn2_gate_param}
\end{align}
The log-decay follows the Gated DeltaNet parameterization,
\begin{align}
    \boldsymbol{g}_t
    = -\exp(\mathbf{a}) \odot \operatorname{softplus}(\mathbf{W}_f\vx_t + \boldsymbol{\delta}),
    \qquad
    \valpha_t = \exp(\boldsymbol{g}_t) .
\label{eq_gdn2_decay_param}
\end{align}
In practice this decay activation is computed in fp32 before the kernel consumes it, which avoids precision loss in the cumulative log-decay. We also support the negative-eigenvalue variant of~\citet{Grazzi2024UnlockingSI} by scaling only the erase gate to $[0,2]^{d_k}$. The write gate remains in $[0,1]^{d_v}$ because the spectral effect concerns the state transition, not the value magnitude.

\subsection{Fast-weight update perspective}

We can interpret Gated Delta Rule-2 as an online update of a fast-weight memory state~\citep{longhorn}. The state $\rmS_t$ stores transient key-value associations. At each token, the model first forms a decayed state $\bar{\rmS}_t=\rmD_t\rmS_{t-1}$, reads the old content through the gated erase direction $\ve_t$, and writes a correction toward the gated value target $\vz_t$.

More formally, Eq.~\ref{eq_gdn2_residual} is the solution of the local online problem
\begin{align}
    \rmS_t
    =
    \operatorname*{arg\,min}_{\rmS}
    \boldsymbol{L}_t(\rmS),
    \qquad
    \boldsymbol{L}_t(\rmS)
    =
    \|\rmS-\bar{\rmS}_t\|_F^2
    -
    2
    \left\langle
    \rmS^\top\vk_t,
    \vz_t-\bar{\rmS}_t^\top\ve_t
    \right\rangle .
\label{eq_gdn2_online_objective}
\end{align}
The first term keeps the new state close to the decayed memory. The second term applies an associative edit whose residual compares the gated write target $\vz_t$ against the content read from $\bar{\rmS}_t$ along $\ve_t$. Since
\begin{align}
    \nabla_{\rmS}\boldsymbol{L}_t(\rmS)
    =
    2(\rmS-\bar{\rmS}_t)
    -
    2\vk_t
    \left(
    \vz_t-\bar{\rmS}_t^\top\ve_t
    \right)^\top ,
\end{align}
the minimizer is
\begin{align}
    \rmS_t
    =
    \bar{\rmS}_t
    +
    \vk_t
    \left(
    \vz_t-\bar{\rmS}_t^\top\ve_t
    \right)^\top ,
\end{align}
which is exactly Eq.~\ref{eq_gdn2_residual}.

Table~\ref{tab_gdn2_online_learning} compares this view with Mamba-2, Gated DeltaNet, KDA, and Mamba-3. We write all updates in the state orientation used in this paper, where $\vo_t=\rmS_t^\top\vq_t$. Normalizer terms, kernel maps, output gates, and value projection gates are omitted for readability.

For the Mamba-3 row, we use the SISO exponential-trapezoidal recurrence~\citep{lahoti2026mamba3}. Let
\begin{align}
    \widetilde{\vk}_s &= \rmR_{1:s}^\top\vk_s,
    &
    \eta_t &= (1-\lambda_t)\Delta_t\alpha_t,
    &
    \zeta_t &= \lambda_t\Delta_t .
\end{align}
Here $\rmR_{1:s}$ is the cumulative data-dependent rotation from the complex SSM view, and the previous-token term is omitted at the beginning of a sequence. The MIMO version replaces each rank-one write with a sum over the MIMO rank and leaves the same online form intact.

\begin{table*}[t]
\caption{
Fast-weight update view of DeltaNet, Mamba-2, Gated DeltaNet, KDA, Mamba-3, and Gated DeltaNet-2.
All updates use the state orientation of this paper, where $\vo_t=\rmS_t^\top\vq_t$.
Mamba-2 and Mamba-3 add gated key-value correlation terms to a decayed state.
DeltaNet, Gated DeltaNet, KDA, and Gated DeltaNet-2 instead write a delta residual, the target value minus the value currently read from memory.
}
\scriptsize
\setlength{\tabcolsep}{3pt}
\renewcommand{\arraystretch}{1.35}
\begin{tabular*}{\linewidth}{@{}l@{\hspace{0.018\linewidth}}p{0.455\linewidth}@{\hspace{0.018\linewidth}}p{0.355\linewidth}@{}}
\toprule
\textbf{Method}
&
\textbf{Local objective $\boldsymbol{L}_t(\rmS)$}
&
\textbf{State update}
\\
\midrule

DeltaNet~\citep{yang2024parallelizing}
&
\(\displaystyle
\begin{aligned}[t]
&\|\rmS-\rmS_{t-1}\|_F^2
\\
&-
2
\left\langle
\rmS^\top\vk_t,
\beta_t
\left(
\vv_t-\rmS_{t-1}^\top\vk_t
\right)
\right\rangle
\end{aligned}
\)
&
\(\displaystyle
\begin{aligned}[t]
\rmS_t
&=
(\rmI-\beta_t\vk_t\vk_t^\top)\rmS_{t-1}
\\
&\quad+
\beta_t\vk_t\vv_t^\top
\end{aligned}
\)
\\
\addlinespace[0.45em]

Mamba-2~\citep{pmlr-v235-dao24a}
&
\(\displaystyle
\begin{aligned}[t]
&\|\rmS-\alpha_t\rmS_{t-1}\|_F^2
\\
&-
2
\left\langle
\rmS^\top\vk_t,
\vv_t
\right\rangle
\end{aligned}
\)
&
\(\displaystyle
\begin{aligned}[t]
\rmS_t
&=
\alpha_t\rmS_{t-1}
+
\vk_t\vv_t^\top
\end{aligned}
\)
\\
\addlinespace[0.45em]

Gated DeltaNet~\citep{yang2025gated}
&
\(\displaystyle
\begin{aligned}[t]
&\|\rmS-\alpha_t\rmS_{t-1}\|_F^2
\\
&-
2
\left\langle
\rmS^\top\vk_t,
\beta_t
\left(
\vv_t-(\alpha_t\rmS_{t-1})^\top\vk_t
\right)
\right\rangle
\end{aligned}
\)
&
\(\displaystyle
\begin{aligned}[t]
\rmS_t
&=
\alpha_t
(\rmI-\beta_t\vk_t\vk_t^\top)
\rmS_{t-1}
\\
&\quad+
\beta_t\vk_t\vv_t^\top
\end{aligned}
\)
\\
\addlinespace[0.45em]

KDA~\citep{team2025kimi}
&
\(\displaystyle
\begin{aligned}[t]
&\|\rmS-\rmD_t\rmS_{t-1}\|_F^2
\\
&-
2
\left\langle
\rmS^\top\vk_t,
\beta_t
\left(
\vv_t-(\rmD_t\rmS_{t-1})^\top\vk_t
\right)
\right\rangle
\end{aligned}
\)
&
\(\displaystyle
\begin{aligned}[t]
\rmS_t
&=
(\rmI-\beta_t\vk_t\vk_t^\top)
\rmD_t\rmS_{t-1}
\\
&\quad+
\beta_t\vk_t\vv_t^\top
\end{aligned}
\)
\\
\addlinespace[0.45em]

Mamba-3~\citep{lahoti2026mamba3}
&
\(\displaystyle
\begin{aligned}[t]
&\|\rmS-\alpha_t\rmS_{t-1}\|_F^2
\\
&-
2
\left\langle
\rmS^\top\widetilde{\vk}_{t-1},
\eta_t\vv_{t-1}
\right\rangle
\\
&-
2
\left\langle
\rmS^\top\widetilde{\vk}_{t},
\zeta_t\vv_t
\right\rangle
\end{aligned}
\)
&
\(\displaystyle
\begin{aligned}[t]
\rmS_t
&=
\alpha_t\rmS_{t-1}
+
\eta_t\widetilde{\vk}_{t-1}\vv_{t-1}^\top
\\
&\quad+
\zeta_t\widetilde{\vk}_{t}\vv_t^\top
\end{aligned}
\)
\\
\addlinespace[0.45em]

Gated DeltaNet-2
&
\(\displaystyle
\begin{aligned}[t]
&\|\rmS-\rmD_t\rmS_{t-1}\|_F^2
\\
&-
2
\left\langle
\rmS^\top\vk_t,
\vz_t-(\rmD_t\rmS_{t-1})^\top\ve_t
\right\rangle
\end{aligned}
\)
&
\(\displaystyle
\begin{aligned}[t]
\rmS_t
&=
(\rmI-\vk_t\ve_t^\top)
\rmD_t\rmS_{t-1}
\\
&\quad+
\vk_t\vz_t^\top
\end{aligned}
\)
\\

\bottomrule
\end{tabular*}
\label{tab_gdn2_online_learning}
\end{table*}

The comparison separates two families. Mamba-2 and Mamba-3 write correlations into a decayed state. Mamba-3 makes this write more expressive through the exponential-trapezoidal input rule and data-dependent rotations, but it does not subtract a current read from the state. Gated DeltaNet and KDA instead perform a residual delta edit. KDA changes the decay from scalar to channel-wise while keeping the scalar residual $\beta_t(\vv_t-\bar{\rmS}_t^\top\vk_t)$. Gated DeltaNet-2 changes the residual itself to
\begin{align}
    \vz_t-\bar{\rmS}_t^\top\ve_t
    =
    \vw_t\odot\vv_t
    -
    (\rmD_t\rmS_{t-1})^\top(\vb_t\odot\vk_t),
\end{align}
which decouples the coordinates used to erase from the coordinates used to write.

\subsection{Chunkwise parallel training}
\label{subsec_chunkwise_training}

We now show that Gated Delta Rule-2 keeps the same chunkwise structure as KDA. Consider one chunk and suppress the chunk index. Let
\begin{align}
    \boldsymbol{G}_r = \sum_{i=1}^{r}\boldsymbol{g}_i,
    \qquad
    \boldsymbol{\gamma}_r = \exp(\boldsymbol{G}_r),
    \qquad
    \boldsymbol{\gamma}_0 = \mathbf{1}_{d_k} .
\label{eq_gdn2_gamma}
\end{align}
Define the decay-normalized state $\widehat{\rmS}_r$ by $\rmS_r = \operatorname{Diag}(\boldsymbol{\gamma}_r)\widehat{\rmS}_r$. With $\widehat{\rmS}_0=\rmS_{[n]}$, Eq.~\ref{eq_gdn2_recurrence} becomes a pure asymmetric delta recurrence,
\begin{align}
    \widehat{\rmS}_r
    = \bigl(\rmI - \bar{\vk}_r\bar{\ve}_r^\top\bigr)\widehat{\rmS}_{r-1}
    + \bar{\vk}_r\vz_r^\top,
    \qquad
    \bar{\vk}_r = \boldsymbol{\gamma}_r^{-1}\odot\vk_r,
    \qquad
    \bar{\ve}_r = \boldsymbol{\gamma}_r\odot\ve_r .
\label{eq_gdn2_normalized_delta}
\end{align}
This normalization is the key to the efficient form. The channel-wise decay is absorbed into the two factors of each rank-one erase, while the update remains a product of matrices of the form $\rmI - \bar{\vk}_r\bar{\ve}_r^\top$.

Let $\rmB\in\mathbb{R}^{C\times d_k}$ and $\rmW\in\mathbb{R}^{C\times d_v}$ contain rows $\vb_r^\top$ and $\vw_r^\top$, respectively. For compact matrix notation, let $\boldsymbol{\gamma}\in\mathbb{R}^{C\times d_k}$ contain rows $\boldsymbol{\gamma}_r^\top$. Let $\bar{\rmK}\in\mathbb{R}^{C\times d_k}$, $\bar{\rmE}\in\mathbb{R}^{C\times d_k}$, and $\rmZ\in\mathbb{R}^{C\times d_v}$ contain rows $\bar{\vk}_r^\top$, $\bar{\ve}_r^\top$, and $\vz_r^\top$. Equivalently,
\begin{align}
    \bar{\rmE}
    =
    \boldsymbol{\gamma}\odot(\rmB\odot\rmK),
    \qquad
    \rmZ
    =
    \rmW\odot\rmV .
\label{eq_gdn2_gate_matrices}
\end{align}
Define the strictly lower triangular matrix
\begin{align}
    \rmT = \operatorname{tril}(\bar{\rmE}\bar{\rmK}^\top, -1),
    \qquad
    \rmA = (\rmI + \rmT)^{-1} .
\label{eq_gdn2_wy_matrix}
\end{align}
The WY auxiliaries are
\begin{align}
    \rmY = \rmA\bar{\rmE},
    \qquad
    \rmU = \rmA\rmZ .
\label{eq_gdn2_wy_aux}
\end{align}
Here $\rmY$ is the erase-side auxiliary and $\rmU$ is the write-side auxiliary. Since $\rmT$ is triangular with zero diagonal, $\rmA$ is obtained by a small forward substitution inside each chunk.

The end-of-chunk state is then
\begin{align}
    \rmS_{[n+1]}
    =
    \operatorname{Diag}(\boldsymbol{\gamma}_C)\rmS_{[n]}
    +
    \rmK_{\mathrm{tail}}^\top(\rmU - \rmY\rmS_{[n]}),
\label{eq_gdn2_chunk_state}
\end{align}
where row $r$ of $\rmK_{\mathrm{tail}}$ is $(\boldsymbol{\gamma}_C / \boldsymbol{\gamma}_r)\odot\vk_r$. The output block is
\begin{align}
    \rmO_{[n]}
    =
    \rmQ_{\gamma}\rmS_{[n]}
    +
    \rmA_{qk}(\rmU - \rmY\rmS_{[n]}),
\label{eq_gdn2_chunk_output}
\end{align}
where row $r$ of $\rmQ_{\gamma}$ is $\boldsymbol{\gamma}_r\odot\vq_r$ and
\begin{align}
    (\rmA_{qk})_{rs}
    =
    \mathbf{1}_{r\ge s}\,
    \vq_r^\top
    \operatorname{Diag}(\boldsymbol{\gamma}_r / \boldsymbol{\gamma}_s)
    \vk_s .
\label{eq_gdn2_chunk_scores}
\end{align}
Equations~\ref{eq_gdn2_chunk_state} and~\ref{eq_gdn2_chunk_output} have the same shape as the KDA chunk equations. The only difference is how $\rmY$ and $\rmU$ are formed. The erase gate enters through row $r$ of $\bar{\rmE}$ as $\boldsymbol{\gamma}_r\odot(\vb_r\odot\vk_r)$. The write gate enters through row $r$ of $\rmZ$ as $\vw_r\odot\vv_r$. The rest of the computation is a triangular solve and dense matrix multiplication over fixed-size chunks. We use the UT transform~\citep{Joffrain2006AccumulatingHT} and implement these equations with fused Triton kernels~\citep{tillet_triton_2019}. Kernel schedules and precision choices are deferred to the supplement.

\subsection{Gate-aware backward}
\label{subsec_gate_aware_backward}

The backward pass follows the same decomposition as the forward. Gradients first flow through the output equation and the inter-chunk state recurrence, both of which operate only on $\rmA_{qk}$, $\rmK_{\mathrm{tail}}$, $\rmY$, and $\rmU$. The only new accounting is the vector-Jacobian product through Eq.~\ref{eq_gdn2_wy_aux} and Eq.~\ref{eq_gdn2_wy_matrix}.

For scalar-gated delta rules, a factor $\beta_r$ can be moved outside the dot products that accumulate the gradient of $\rmA$. That shortcut breaks for Gated Delta Rule-2. The write side contains a different diagonal gate over value channels, and the erase side contains a different diagonal gate over key channels. Therefore the gate factors must be present at the accumulation sites. Let $\rmB$ and $\rmW$ contain rows $\vb_r^\top$ and $\vw_r^\top$, respectively, and let $\boldsymbol{\gamma}$ denote the row-stacked cumulative-decay vectors. Then
\begin{align}
    \mathrm{d}\rmA &\mathrel{+}= \mathrm{d}\rmU\,\rmZ^\top,
    &
    \rmZ &= \rmW\odot\rmV,
\label{eq_gdn2_backward_u}\\
    \mathrm{d}\rmA &\mathrel{+}= \mathrm{d}\rmY\,\bar{\rmE}^\top,
    &
    \bar{\rmE} &= \boldsymbol{\gamma}\odot(\rmB\odot\rmK) .
\label{eq_gdn2_backward_y}
\end{align}
The inverse itself has the standard triangular vector-Jacobian product
\begin{align}
    \mathrm{d}\rmT =
    -\operatorname{tril}
    \bigl(
    \rmA^\top\mathrm{d}\rmA\rmA^\top,
    -1
    \bigr).
\label{eq_gdn2_inverse_vjp}
\end{align}
From there, gradients to $\rmB$, $\rmW$, $\rmK$, $\rmV$, and the cumulative decay follow by ordinary elementwise products and reverse cumulative sums. This gate-aware accumulation is the main mathematical change required for training Gated Delta Rule-2. The remaining backward kernels retain the same matrix shapes as KDA and can reuse the same state and output vector-Jacobian product structure.

\subsection{Block design and hybrid models}
\label{subsec_block_design}

\definecolor{fgate_color}{RGB}{252,224,225}
\definecolor{wgate_color}{RGB}{255,210,210}
\definecolor{delta_color}{RGB}{242,243,193}
\definecolor{swa_color}{RGB}{252,224,225}
\definecolor{add_norm_color}{RGB}{252,226,187}
\definecolor{glu_color}{RGB}{194,232,247}
\definecolor{silu_color}{RGB}{203,231,207}
\definecolor{linear_color}{RGB}{220,223,240}
\definecolor{conv_color}{RGB}{252,224,225}
\definecolor{l2_color}{RGB}{252,226,187}
\definecolor{gray_bbox_color}{RGB}{243,243,244}
\definecolor{oproj_color}{RGB}{220,223,240}
\definecolor{operator_color}{RGB}{252,224,225}

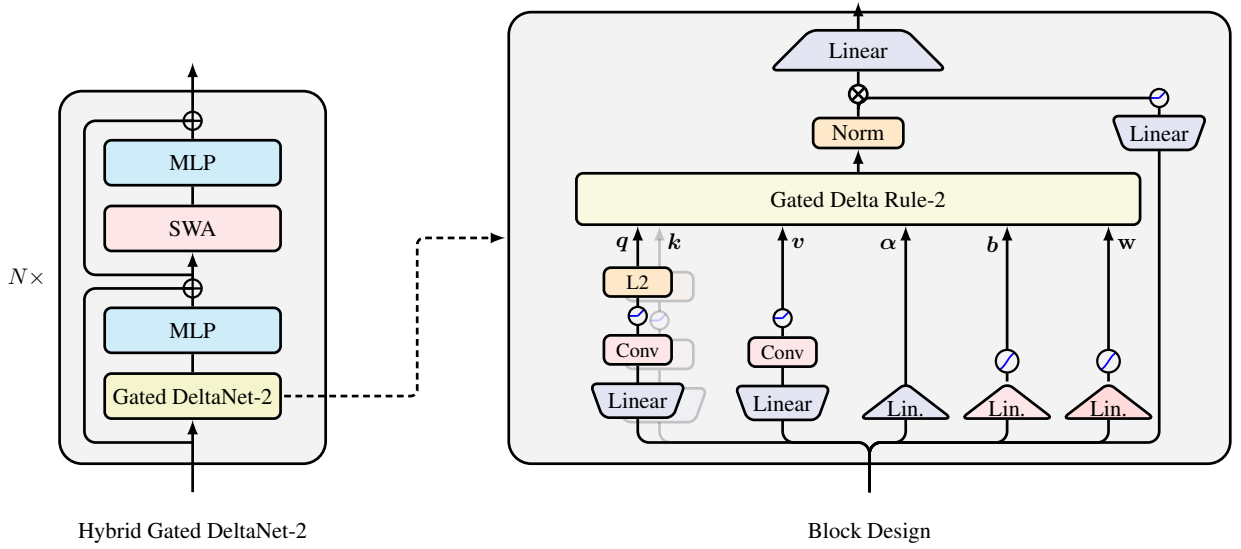
\begin{figure}
\centering
\scalebox{0.99}{
\tikzset{
model/.style={
    draw=black, very thick, fill=gray_bbox_color,
    minimum width=118pt, rounded corners=10pt
  },
gdelta/.style={
    draw=black, very thick, fill=gray_bbox_color,
    minimum width=200pt, minimum height=200pt, rounded corners=10pt
  },
tokenmixer/.style={
    draw=black, very thick, fill=delta_color!80,
    minimum width=78pt, minimum height=0.7cm, rounded corners=3pt
  },
swa/.style={
    draw=black, very thick, fill=swa_color!80,
    minimum width=78pt, minimum height=0.7cm, rounded corners=3pt
  },
glu/.style={
    draw=black, very thick, fill=glu_color!80,
    minimum width=78pt, minimum height=0.7cm, rounded corners=3pt
  },
norm/.style={
    draw=black, very thick, fill=add_norm_color!80,
    minimum width=40pt, rounded corners=3pt, align=center,
  },
linear/.style={
    draw=black, very thick, fill=oproj_color!80,
    minimum width=40pt, rounded corners=3pt
  },
stacked/.style={
    draw=black, very thick, fill=linear_color!80,
    minimum width=40pt, minimum height=15pt, rounded corners=3pt,
  },
trapezoid/.style={
    trapezium, trapezium left angle=110, trapezium right angle=110,
    rounded corners=3pt, inner xsep=1pt, outer sep=0pt,
  },
conv/.style={
    draw=black, very thick, minimum width=30pt,
    fill=conv_color!80, rounded corners=3pt, rectangle, font=\small
  },
l2/.style={
    draw=black, very thick, minimum width=30pt,
    fill=l2_color!80, rounded corners=3pt, rectangle, font=\small
  },
fgate/.style={
    draw=black, very thick, fill=fgate_color!80,
    minimum width=30pt, minimum height=15pt, rounded corners=3pt
  },
oproj/.style={
    draw=black, very thick, fill=oproj_color!80,
    minimum width=78pt, rounded corners=3pt
  },
silu/.style={
    draw=black, very thick, fill=silu_color,
    minimum width=1.1cm, rounded corners=3pt
  },
layerlink/.style={ -latex, very thick, },
modulelink/.style={
    -latex, very thick, densely dashed,
    shorten >=1pt, shorten <=1pt, rounded corners=3pt
  },
normlink/.style={ very thick, },
residual/.style={ very thick, rounded corners=5pt },
qf/.style={ -latex, very thick, rounded corners=5pt },
oplus/.style={
    draw=black, line width=1pt, circle, minimum size=8pt,
    inner sep=0pt, outer sep=0pt,
    path picture={
        \draw (path picture bounding box.center) -- ++(0.3cm,0)
        (path picture bounding box.center) -- ++(-0.3cm,0)
        (path picture bounding box.center) -- ++(0,0.3cm)
        (path picture bounding box.center) -- ++(0,-0.3cm);
      },
  },
otimes/.style={
    draw=black, very thick, circle, minimum size=8pt,
    inner sep=0pt, outer sep=0pt,
    path picture={
        \draw (path picture bounding box.center) -- ++(0.25cm,0.25cm)
        (path picture bounding box.center) -- ++(-0.25cm,-0.25cm)
        (path picture bounding box.center) -- ++(-0.25cm,0.25cm)
        (path picture bounding box.center) -- ++(0.25cm,-0.25cm);
      }
  },
sigmoid/.style={
    draw=black, thick, line width=1pt, circle, minimum size=10pt,
    inner sep=0pt, outer sep=0pt,
    path picture={
        \draw[domain=-2:2, samples=40, variable=\x, blue, thick]
        plot ({\x*0.13}, {0.32/(1+exp(-2.5*\x)) - 0.16});
      }
  },
swish/.style={
    draw=black, thick, line width=1pt, circle, minimum size=8pt,
    inner sep=0pt, outer sep=0pt,
    path picture={
        \draw[domain=-1.5:0, samples=50, variable=\x, blue, thick]
        plot ({\x}, {0});
        \draw[domain=0:1.5, samples=50, variable=\x, blue, thick]
        plot ({\x}, {\x});
      }
  },
kgdelta/.style={
    draw=black, very thick, fill=delta_color!50,
    minimum width=115pt, minimum height=0.8cm, rounded corners=3pt
  },
}
\resizebox{\textwidth}{!}{
\begin{tikzpicture}
\centering


\node[model, minimum height=165pt] (model) at (0,0) {};
\node[anchor=east,xshift=-2pt] at (model.west) (ntimes) {$N\times$};

\node[tokenmixer, anchor=south, yshift=20pt] at (model.south) (gdelta1) {Gated DeltaNet-2};
\draw[layerlink] ($(model.south) + (0,-12pt)$) -- (gdelta1.south);

\node[glu, anchor=south, yshift=8pt] at (gdelta1.north) (mlp1) {MLP};
\draw[normlink] (gdelta1.north) -- (mlp1.south);

\node[oplus, anchor=south, yshift=4pt] at (mlp1.north) (oplus1) {};
\draw[normlink] (mlp1.north) -- (oplus1.south);

\node[swa, anchor=south, yshift=12pt] at (oplus1.north) (swa) {SWA};
\draw[layerlink] (oplus1.north) -- (swa.south);

\draw[residual] ([yshift=-10pt]gdelta1.south) -- ([xshift=-48pt,yshift=-10pt]gdelta1.south) -- ([xshift=-48pt]oplus1.center) -- (oplus1.center);

\node[glu, anchor=south, yshift=8pt] at (swa.north) (mlp2) {MLP};
\draw[normlink] (swa.north) -- (mlp2.south);

\node[oplus, anchor=south, yshift=4pt] at (mlp2.north) (oplus2) {};
\draw[normlink] (mlp2.north) -- (oplus2.south);

\draw[residual] ([xshift=0pt,yshift=6pt]oplus1.center) -- ([xshift=-48pt,yshift=6pt]oplus1.center) -- ([xshift=-48pt]oplus2.center) -- (oplus2.center);

\node[above=12pt] at (model.north) (modeloutput) {\textcolor{white}{Outputs}};
\draw[layerlink] (oplus2.north) -- (modeloutput.south);

\node[below=22pt] at (model.south) (input1) {Hybrid Gated DeltaNet-2};


\node[gdelta, anchor=south west, xshift=80pt, minimum width=320pt] (gdelta) at (model.south east) {};
\node[] at (gdelta |- input1) (input2) {Block Design};

\node[kgdelta, xshift=-5pt, yshift=15pt, minimum width=250pt] at (gdelta.mid) (kernel) {Gated Delta Rule-2};
\node[below=12pt] at (gdelta.south) (input) {\textcolor{white}{Inputs}};

\node[stacked, trapezium, trapezium left angle=110, trapezium right angle=110, inner xsep=1pt, outer sep=0pt, minimum width=30pt, anchor=south, yshift=20pt] at ($(gdelta.south west)!0.38!(gdelta.south east)$) (vproj) {Linear};
\node[conv, anchor=south, yshift=8pt] at (vproj.north) (vconv) {Conv};
\node[swish, anchor=south, yshift=4pt] at (vconv.north) (vsilu) {};
\draw[layerlink] (vsilu.north) -- (vsilu|-kernel.south) node[pos=0.8, right] {$\vv$};
\draw[normlink] (vsilu.south) -- (vconv.north);
\draw[normlink] (vconv.south) -- (vproj.north);
\draw[residual] (input.north) -- (gdelta.south) |- ([xshift=-10pt,yshift=10pt]gdelta.south) -| (vproj.south);

\node[stacked, trapezium, trapezium left angle=110, trapezium right angle=110, inner xsep=1pt, outer sep=0pt, minimum width=30pt, anchor=south, yshift=19pt, opacity=0.2] at ($(gdelta.south west)!0.21!(gdelta.south east)$) (kproj) {\textcolor{white}{Linear}};
\node[conv, anchor=south, yshift=8pt, opacity=0.2] at (kproj.north) (kconv) {\textcolor{white}{Conv}};
\node[swish, anchor=south, yshift=4pt, opacity=0.2] at (kconv.north) (ksilu) {};
\node[l2, anchor=south, yshift=4pt, opacity=0.2] at (ksilu.north) (kl2) {\textcolor{white}{L2}};
\draw[layerlink, opacity=0.2] (kl2.north) -- (kl2|-kernel.south);
\node[right] at ($(kl2)+(0,20pt)$) {$\vk$};
\draw[normlink, opacity=0.2] (ksilu.north) -- (kl2.south);
\draw[normlink, opacity=0.2] (ksilu.south) -- (kconv.north);
\draw[normlink, opacity=0.2] (kconv.south) -- (kproj.north);
\draw[residual, opacity=0.2] (input.north) -- (gdelta.south) |- ([xshift=-10pt,yshift=10pt]gdelta.south) -| (kproj.south);

\node[stacked, trapezium, trapezium left angle=110, trapezium right angle=110, inner xsep=1pt, outer sep=0pt, minimum width=30pt, anchor=south, yshift=21pt] at ($(gdelta.south west)!0.18!(gdelta.south east)$) (qproj) {Linear};
\node[conv, anchor=south, yshift=8pt] at (qproj.north) (qconv) {Conv};
\node[swish, anchor=south, yshift=4pt] at (qconv.north) (qsilu) {};
\node[l2, anchor=south, yshift=4pt] at (qsilu.north) (ql2) {L2};
\draw[layerlink] (ql2) -- (ql2|-kernel.south) node[pos=0.55, left] {$\vq$};
\draw[normlink] (qsilu.north) -- (ql2.south);
\draw[normlink] (qsilu.south) -- (qconv.north);
\draw[normlink] (qconv.south) -- (qproj.north);
\draw[residual] (input.north) -- (gdelta.south) |- ([xshift=-10pt,yshift=10pt]gdelta.south) -| (qproj.south);

\node[stacked, isosceles triangle, isosceles triangle apex angle=105, draw=black, very thick, inner sep=.9pt, anchor=south, yshift=20pt, shape border rotate=90] at ($(gdelta.south west)!0.55!(gdelta.south east)$) (fproj) {Lin.};
\draw[qf] ($(fproj.north)-(0,2pt)$) -- (fproj.north|-kernel.south) node[pos=0.89, left] {$\valpha$};
\draw[residual] (input.north) -- (gdelta.south) |- ([xshift=10pt,yshift=10pt]gdelta.south) -| (fproj.south);

\node[stacked, isosceles triangle, isosceles triangle apex angle=105, draw=black, very thick, fill=fgate_color!80, inner sep=.9pt, anchor=south, yshift=20pt, shape border rotate=90] at ($(gdelta.south west)!0.69!(gdelta.south east)$) (bproj) {Lin.};
\node[sigmoid, anchor=south, yshift=4pt] at (bproj.north) (bsigmoid) {};
\draw[normlink] (bproj.north) -- (bsigmoid.south);
\draw[qf] (bsigmoid.north) -- (bsigmoid.north|-kernel.south) node[pos=0.85, left] {$\vb$};
\draw[residual] (input.north) -- (gdelta.south) |- ([xshift=10pt,yshift=10pt]gdelta.south) -| (bproj.south);

\node[stacked, isosceles triangle, isosceles triangle apex angle=105, draw=black, very thick, fill=wgate_color!80, inner sep=.9pt, anchor=south, yshift=20pt, shape border rotate=90] at ($(gdelta.south west)!0.83!(gdelta.south east)$) (wproj) {Lin.};
\node[sigmoid, anchor=south, yshift=4pt] at (wproj.north) (wsigmoid) {};
\draw[normlink] (wproj.north) -- (wsigmoid.south);
\draw[qf] (wsigmoid.north) -- (wsigmoid.north|-kernel.south) node[pos=0.85, right] {$\vw$};
\draw[residual] (input.north) -- (gdelta.south) |- ([xshift=10pt,yshift=10pt]gdelta.south) -| (wproj.south);

\node[norm, anchor=south, yshift=10pt] at (kernel.north) (norm4) {Norm};
\node[otimes, anchor=south, yshift=6pt] at (norm4.north) (otimes) {};
\node[draw=black, very thick, fill=oproj_color!80, minimum width=78pt, minimum height=10pt, shape=trapezium, trapezium angle=45, rounded corners=3pt, anchor=south, yshift=6pt] at (otimes.north) (oproj) {Linear};
\draw[layerlink] (kernel.north) -- (norm4.south);
\draw[normlink] (norm4.north) -- (otimes.south);
\draw[normlink] (otimes.north) -- (oproj.south);

\coordinate (gproj_x) at ($(gdelta.south west)!0.90!(gdelta.south east)$);
\node[stacked, trapezium, trapezium left angle=110, trapezium right angle=110, inner xsep=1pt, outer sep=0pt, minimum width=30pt, anchor=center] at (gproj_x |- norm4.center) (gproj) {Linear};
\node[swish, anchor=south, yshift=4pt] at (gproj.north) (osilu) {};
\draw[normlink] (gproj.north) -- (osilu.south);
\draw[residual] (input.north) -- (gdelta.south) |- ([xshift=10pt,yshift=10pt]gdelta.south) -| (gproj.south);
\draw[residual] (osilu) -| (otimes);

\node[above=12pt] at (oproj.north) (blockoutput) {\textcolor{white}{Outputs}};
\draw[layerlink] (oproj.north) -- (blockoutput.south);

\draw[modulelink] (gdelta1.east) -| ([yshift=-10pt]$(model.east)!0.5!(gdelta.west)$) |- (gdelta.west);

\end{tikzpicture}
}
}
\caption{\small
Visualization of the hybrid architecture and block design of \textbf{Gated DeltaNet-2}.
The \emph{Hybrid Gated DeltaNet-2} model repeats a Gated DeltaNet-2 token mixer, an MLP, sliding-window attention (SWA), and another MLP.
In the block design, query and key paths use linear projection, short convolution, SiLU, and L2 normalization.
The value path uses linear projection, short convolution, and SiLU.
The central recurrent operator is \emph{Gated Delta Rule-2}.
The decay branch produces $\valpha$ from the log-decay projection.
The channel-wise erase gate $\vb$ and channel-wise write gate $\vw$ each use linear projection followed by sigmoid.
The recurrent output is normalized, multiplied by a SiLU output gate, and passed through the output projection.
}
\label{fig:gated_deltanet2_model}
\end{figure}

\paragraph{Gated DeltaNet-2 token mixer.}
Gated DeltaNet-2 is used as the recurrent token mixer in a standard Transformer-style block. Fig.~\ref{fig:gated_deltanet2_model} (right) shows its block design. For the Gated Delta Rule-2 in Eq.~\ref{eq_gdn2_recurrence}, $\{\vq_t,\vk_t,\vv_t\}$ are produced by linear projection, short causal convolution, and SiLU, with L2 normalization applied to $\vq_t$ and $\vk_t$ for stability. Separate branches produce the channel-wise decay $\valpha_t$, erase gate $\vb_t$, and write gate $\vw_t$. The recurrent output is RMS-normalized, multiplied by a separate SiLU output gate, and projected back to the model dimension. Throughout the paper, $\boldsymbol{g}$ denotes the log-decay tensor in Eq.~\ref{eq_gdn2_decay_param}, not the output gate. With grouped value heads, $\vq$, $\vk$, the log-decay tensor $\boldsymbol{g}$, and $\vb$ are repeated across value-head groups, while $\vv$ and $\vw$ remain on the value-head axis.

\paragraph{Model families.}
We train both recurrent and hybrid models. The recurrent model stacks Gated DeltaNet-2 token mixers and MLPs under the standard residual block, isolating the fixed-state memory of Eq.~\ref{eq_gdn2_recurrence}. The hybrid model inserts Sliding-Window Attention (SWA) after the recurrent mixer, as shown in Fig.~\ref{fig:gated_deltanet2_model} (left). A repeated cell contains Gated DeltaNet-2, an MLP, SWA, and another MLP. Gated DeltaNet-2 compresses long histories into constant-size memory, while SWA handles exact local interactions such as short shifts, comparisons, and local retrieval. With a fixed window, the hybrid retains linear sequence scaling and a bounded attention cache, following the recurrent attention hybrid design pattern \citep{de_griffin_2024,ren2024samba}.

\section{Experiments}
\label{sec_exp}
\paragraph{Setup}
We evaluate each recurrent family in two forms, a recurrent-only model and a hybrid model that pairs the same recurrent token mixer with sliding-window attention as described in Section~\ref{subsec_block_design}. For Mamba-3, we include both SISO and MIMO variants and use MIMO rank $R=4$ following \citet{lahoti2026mamba3}. All models are trained with the same recipe. Unless stated otherwise, each model has 1.3B parameters and is trained on 100B tokens from FineWeb-Edu \citep{penedo2024fineweb}. We use AdamW with peak learning rate $4\times10^{-4}$, weight decay $0.1$, gradient clipping at $1.0$, cosine decay, a 1B-token warm-up, and a global batch size of 0.5M tokens. The training length is 4K tokens, and hybrid models use a 2K sliding-window attention size. Evaluation details are given in the appendix.

\begin{table*}[t!]
\vspace{0mm}
\centering
\scriptsize
\addtolength{\tabcolsep}{-3.2pt}
\begin{tabular}{l|cc|cccccccccc}
\toprule
\textbf{Model} & \textbf{Wiki.} & \textbf{LMB.} & \textbf{LMB.} & \textbf{PIQA} & \textbf{Hella.} & \textbf{Wino.} & \textbf{ARC-e} & \textbf{ARC-c} & \textbf{OBQA} & \textbf{SIQA} & \textbf{BoolQ} & \textbf{Avg.} \\
 & ppl $\downarrow$ & ppl $\downarrow$ & acc $\uparrow$ & acc $\uparrow$ & acc\_n $\uparrow$ & acc $\uparrow$ & acc $\uparrow$ & acc $\uparrow$ & acc $\uparrow$ & acc $\uparrow$ & acc $\uparrow$ & acc $\uparrow$ \\
\midrule
\midrule
\textit{Recurrent models} \\
\hspace{2mm} Mamba-2 & 16.79 & 12.38 & 45.24 & \underline{72.58} & 55.51 & 55.33 & 70.68 & 35.26 & \underline{31.00} & 40.63 & \underline{60.19} & 51.82 \\
\hspace{2mm} Gated DeltaNet & 16.40 & 11.89 & \textbf{49.62} & 72.31 & \underline{56.50} & \underline{56.75} & 68.81 & 35.15 & 30.20 & 40.53 & 58.78 & 52.07 \\
\hspace{2mm} KDA & 16.81 & 11.68 & \underline{48.13} & 72.09 & 55.75 & 55.72 & 70.83 & 35.92 & 30.40 & \underline{40.99} & \textbf{60.67} & 52.28 \\
\hspace{2mm} Mamba-3 (SISO) & \underline{16.30} & 12.99 & 45.06 & 72.31 & 55.58 & 56.20 & 70.45 & 34.56 & \underline{31.00} & \textbf{41.76} & 55.90 & 51.42 \\
\hspace{2mm} Mamba-3 (MIMO) & 16.45 & \underline{11.66} & 47.82 & 72.36 & 56.49 & 55.78 & \underline{72.38} & \underline{38.07} & 30.00 & 40.89 & 57.74 & \underline{52.39} \\
\hspace{2mm} Gated DeltaNet-2 & \textbf{15.90} & \textbf{11.41} & 48.09 & \textbf{72.80} & \textbf{56.84} & \textbf{57.85} & \textbf{72.43} & \textbf{38.23} & \textbf{31.60} & 40.58 & 59.54 & \textbf{53.11} \\
\midrule
\textit{Attention or hybrid models} \\
\hspace{2mm} Transformer & 19.22 & 13.72 & 48.32 & 70.21 & 56.12 & 55.85 & 69.23 & 33.84 & 25.00 & 39.74 & 59.42 & 50.86 \\
\hspace{2mm} Mamba-2 & 17.46 & 11.29 & 48.05 & 71.47 & 57.52 & 56.17 & 70.50 & 34.73 & 29.80 & 40.35 & 59.31 & 51.99 \\
\hspace{2mm} Gated DeltaNet & 16.00 & 10.82 & 48.71 & 70.06 & 57.50 & 56.83 & 70.41 & 35.15 & 30.60 & 40.97 & 60.00 & 52.25 \\
\hspace{2mm} KDA & 16.01 & 10.66 & 49.21 & 71.06 & 56.89 & \underline{57.77} & \underline{71.59} & 35.07 & 30.00 & 40.53 & \underline{62.03} & 52.68 \\
\hspace{2mm} Mamba-3 (SISO) & \textbf{15.54} & \underline{10.65} & 49.19 & 71.01 & \textbf{58.75} & 57.30 & 70.54 & 36.35 & \underline{32.00} & \underline{41.20} & 57.86 & 52.69 \\
\hspace{2mm} Mamba-3 (MIMO) & 15.81 & 10.92 & \underline{49.82} & \underline{71.98} & 58.19 & 57.06 & 70.54 & \textbf{38.48} & 29.40 & 40.99 & 57.98 & \underline{52.72} \\
\hspace{2mm} Gated DeltaNet-2 & \underline{15.62} & \textbf{10.43} & \textbf{50.90} & \textbf{72.20} & \underline{58.46} & \textbf{58.56} & \textbf{71.89} & \underline{36.69} & \textbf{33.00} & \textbf{41.50} & \textbf{62.57} & \textbf{53.97} \\
\bottomrule
\end{tabular}
\addtolength{\tabcolsep}{3.2pt}
\caption{
Performance comparison on language modeling and zero-shot common-sense reasoning.
All accuracy values are reported as percentages.
Avg. is computed over LAMBADA accuracy and the listed reasoning accuracies.
}
\label{tab_commonsense_results}
\end{table*}


\begin{table*}[t!]
\centering
\scriptsize
\setlength{\tabcolsep}{2.0pt}
\renewcommand{\arraystretch}{1.03}

\newcommand{\niahbest}[1]{\textbf{#1}}
\newcommand{\niahsecond}[1]{\underline{#1}}
\newcommand{\modelname}[1]{\hspace{1.2mm}#1}

\resizebox{0.88\textwidth}{!}{%
\begin{tabular}{@{}l@{\hspace{6pt}}cccc@{\hspace{8pt}}cccc@{\hspace{8pt}}ccc@{\hspace{8pt}}ccc@{}}
\toprule
\textbf{Model} &
\multicolumn{4}{c}{\textbf{S-NIAH-1}} &
\multicolumn{4}{c}{\textbf{S-NIAH-2}} &
\multicolumn{3}{c}{\textbf{S-NIAH-3}} &
\multicolumn{3}{c}{\textbf{MK-NIAH-1}} \\
\cmidrule(lr){2-5}
\cmidrule(lr){6-9}
\cmidrule(lr){10-12}
\cmidrule(l){13-15}
& \textbf{1K} & \textbf{2K} & \textbf{4K} & \textbf{8K}
& \textbf{1K} & \textbf{2K} & \textbf{4K} & \textbf{8K}
& \textbf{1K} & \textbf{2K} & \textbf{4K}
& \textbf{1K} & \textbf{2K} & \textbf{4K} \\
\midrule

\multicolumn{15}{@{}l}{\textit{Recurrent models}} \\
\addlinespace[0.2ex]

\modelname{Mamba-2}
& \niahbest{100.0} & \niahbest{100.0} & 97.0 & 55.8
& 99.6 & \niahsecond{99.6} & 62.6 & 21.0
& 59.2 & 38.6 & 14.4
& 29.0 & 21.2 & 21.4 \\

\modelname{Gated DeltaNet}
& \niahsecond{99.8} & \niahbest{100.0} & \niahbest{100.0} & \niahsecond{97.6}
& \niahbest{100.0} & \niahbest{100.0} & 87.2 & \niahsecond{32.0}
& \niahsecond{89.8} & 54.2 & \niahbest{60.6}
& \niahsecond{58.0} & 37.0 & 27.8 \\

\modelname{KDA}
& \niahbest{100.0} & \niahbest{100.0} & \niahsecond{99.2} & 70.6
& \niahbest{100.0} & \niahbest{100.0} & \niahsecond{89.0} & 30.6
& 77.4 & 63.2 & 26.2
& 54.0 & \niahsecond{44.2} & \niahsecond{28.0} \\

\modelname{Mamba-3 (SISO)}
& \niahbest{100.0} & 99.0 & 63.4 & 27.8
& \niahsecond{99.8} & 99.0 & 59.4 & 25.2
& 60.2 & 35.6 & 12.2
& 44.8 & 27.4 & 20.2 \\

\modelname{Mamba-3 (MIMO)}
& \niahbest{100.0} & \niahsecond{99.8} & 93.0 & 35.6
& \niahsecond{99.8} & 98.8 & 64.2 & 27.2
& 89.2 & \niahsecond{72.4} & 29.2
& 49.4 & 19.2 & 18.0 \\

\modelname{Gated DeltaNet-2}
& \niahbest{100.0} & \niahbest{100.0} & \niahbest{100.0} & \niahbest{97.8}
& \niahbest{100.0} & \niahbest{100.0} & \niahbest{93.0} & \niahbest{39.2}
& \niahbest{92.0} & \niahbest{89.8} & \niahsecond{31.8}
& \niahbest{72.6} & \niahbest{51.4} & \niahbest{37.8} \\

\midrule

\multicolumn{15}{@{}l}{\textit{Attention or hybrid models}} \\
\addlinespace[0.2ex]

\modelname{Transformer}
& \niahbest{100.0} & \niahbest{100.0} & 51.2 & 0.0
& \niahbest{100.0} & \niahbest{100.0} & 44.2 & 0.0
& 95.8 & 94.8 & 37.0
& 75.6 & 66.6 & 38.2 \\

\modelname{Mamba-2}
& \niahbest{100.0} & \niahbest{100.0} & \niahsecond{51.8} & 25.4
& \niahbest{100.0} & 99.6 & 52.4 & 25.8
& 97.8 & 86.8 & 48.0
& 82.0 & 58.6 & 39.0 \\

\modelname{Gated DeltaNet}
& \niahbest{100.0} & \niahbest{100.0} & 47.2 & 22.4
& \niahbest{100.0} & \niahsecond{99.8} & 57.3 & 25.6
& 94.8 & 91.2 & 47.2
& 91.0 & 78.4 & 44.8 \\

\modelname{KDA}
& \niahbest{100.0} & \niahbest{100.0} & \niahsecond{51.8} & \niahsecond{26.2}
& \niahbest{100.0} & \niahbest{100.0} & 56.0 & 23.0
& 97.2 & 93.4 & 51.6
& \niahsecond{91.4} & \niahsecond{84.0} & 40.4 \\

\modelname{Mamba-3 (SISO)}
& \niahbest{100.0} & \niahbest{100.0} & 49.6 & 26.0
& \niahbest{100.0} & \niahbest{100.0} & \niahbest{58.2} & \niahsecond{27.8}
& 95.0 & 90.4 & 44.0
& 78.8 & 65.6 & 33.6 \\

\modelname{Mamba-3 (MIMO)}
& \niahbest{100.0} & \niahbest{100.0} & 49.0 & 22.8
& \niahbest{100.0} & \niahbest{100.0} & 53.0 & \niahsecond{27.8}
& \niahsecond{99.4} & \niahsecond{98.4} & \niahsecond{54.2}
& 82.4 & 79.0 & \niahsecond{46.6} \\

\modelname{Gated DeltaNet-2}
& \niahbest{100.0} & \niahbest{100.0} & \niahbest{55.2} & \niahbest{27.4}
& \niahbest{100.0} & \niahbest{100.0} & \niahsecond{57.9} & \niahbest{29.2}
& \niahbest{99.6} & \niahbest{99.0} & \niahbest{55.6}
& \niahbest{93.0} & \niahbest{84.6} & \niahbest{48.0} \\

\bottomrule
\end{tabular}
}
\caption{
Accuracy on Single Needle-In-A-Haystack (S-NIAH) and Multi-Key Needle-In-A-Haystack (MK-NIAH) tasks from RULER.
Best values within each model family and context length are bolded; second-best values are underlined.
}
\label{tab_niah_results}
\vspace{-3mm}
\end{table*}
\paragraph{Language modeling and common-sense reasoning}
Table~\ref{tab_commonsense_results} reports WikiText and LAMBADA perplexity \citep{merity2016pointer,paperno_lambada_2016}, zero-shot LAMBADA accuracy, and the common-sense suite from PIQA through BoolQ \citep{bisk2020piqa,zellers2019hellaswag,sakaguchi2021winogrande,arc-ce,openbookqa,sap2019social,clark2019boolq}. Gated DeltaNet-2 achieves the best average in both recurrent and hybrid settings. Since recurrent state size is matched, the gain points to a stronger update rule rather than a larger memory. The trend persists with SWA, and the model is more balanced than Mamba-3 across perplexity, accuracy, and transfer.

\paragraph{In-context retrieval on synthetic data}
Table~\ref{tab_niah_results} reports S-NIAH and MK-NIAH from RULER \citep{hsieh2024ruler}, which test retention, interference control, high-entropy value storage, and multi-key discrimination under fixed-state memory. Gated DeltaNet-2 is strongest where memory editing matters most. In the recurrent setting, it leads the interference-heavy S-NIAH-2 cases at 4K and 8K and all MK-NIAH-1 lengths. The hybrid model shows the same pattern, leading the long S-NIAH-1 cases, the 8K S-NIAH-2 case, all S-NIAH-3 lengths, and the longer MK-NIAH-1 settings. These gains match the design of Gated Delta Rule-2. The key-side erase gate $\vb_t$ selectively protects or revises key channels, while the value-side write gate $\vw_t$ controls which value channels enter the state. With SWA handling local evidence, this decoupled recurrent update preserves longer-range associations more effectively than a scalar delta gate.

\begin{wraptable}[16]{r}{0.62\linewidth}
\vspace{-1.3\baselineskip}
\centering
\scriptsize
\setlength{\tabcolsep}{1.3pt}
\setlength{\abovecaptionskip}{2pt}
\setlength{\belowcaptionskip}{-4pt}
\begin{tabular}{l|ccccccc}
\toprule
\textbf{Models} & \textbf{SWDE} & \textbf{SQD} & \textbf{FDA} & \textbf{TQA} & \textbf{NQ} & \textbf{DROP} & \textbf{Avg.} \\
\midrule
\midrule
\textit{Recurrent models} \\
\hspace{2mm} Mamba-2 & 17.24 & 32.38 & 14.53 & 58.35 & 18.91 & 19.60 & 26.84 \\
\hspace{2mm} Gated DeltaNet & 17.90 & 32.67 & \underline{18.52} & \underline{59.60} & \textbf{20.16} & 19.69 & 28.09 \\
\hspace{2mm} Mamba-3 (SISO) & 17.62 & 35.07 & 11.08 & 58.89 & 18.18 & \underline{21.32} & 27.03 \\
\hspace{2mm} KDA & \underline{22.49} & 35.10 & 14.90 & 58.12 & 19.58 & \textbf{21.80} & \underline{28.67} \\
\hspace{2mm} Mamba-3 (MIMO) & 16.68 & \underline{36.65} & 17.44 & 59.06 & 19.16 & 21.08 & 28.35 \\
\hspace{2mm} Gated DeltaNet-2 & \textbf{23.65} & \textbf{36.75} & \textbf{19.98} & \textbf{61.37} & \underline{19.64} & 17.87 & \textbf{29.88} \\
\midrule
\textit{Attention or hybrid models} \\
\hspace{2mm} Transformer & 32.21 & 38.67 & 54.78 & 58.09 & 22.49 & 22.18 & 38.07 \\
\hspace{2mm} Mamba-2 & 34.67 & 40.74 & 52.31 & 60.13 & 25.91 & \textbf{24.68} & 39.74 \\
\hspace{2mm} Gated DeltaNet & 33.18 & 42.28 & 50.86 & \underline{60.60} & 25.78 & 21.95 & 39.11 \\
\hspace{2mm} Mamba-3 (SISO) & 35.30 & \textbf{46.42} & \underline{54.95} & 59.54 & 25.91 & \underline{23.96} & \underline{41.01} \\
\hspace{2mm} KDA & \underline{39.83} & 40.10 & 53.59 & 59.89 & 25.27 & 22.18 & 40.14 \\
\hspace{2mm} Mamba-3 (MIMO) & 32.33 & \underline{44.70} & \textbf{55.31} & 59.00 & \underline{26.26} & 23.08 & 40.11 \\
\hspace{2mm} Gated DeltaNet-2 & \textbf{41.96} & \underline{44.70} & 54.68 & \textbf{62.38} & \textbf{26.31} & 23.67 & \textbf{42.28} \\
\bottomrule
\end{tabular}
\caption{
Accuracy on real-world retrieval tasks with input length truncated to 2K tokens.
SQD denotes SQuAD. TQA denotes TriviaQA.
}
\label{tab_recall_results}
\vspace{-1\baselineskip}
\end{wraptable}
\paragraph{In-context retrieval on real-world tasks}
Table~\ref{tab_recall_results} reports recall-heavy real-world tasks from \citet{arora-2024-jrt}, spanning extraction, question answering, and distractor-rich evidence. These tasks are less controlled than synthetic NIAH but better reflect fixed-state memory under realistic context. Gated DeltaNet-2 achieves the best average in both recurrent and hybrid settings. Its recurrent gains are strongest on noisy association recovery, where selective erase and gated write are directly useful. The remaining NQ and DROP gaps point to formats that also need local evidence aggregation, which SWA supplies in the hybrid model.

\begin{table*}[t]
\centering
\scriptsize
\setlength{\tabcolsep}{1.5pt}
\renewcommand{\arraystretch}{1.08}
\begin{tabular}{@{}l@{\hspace{5pt}}cc@{\hspace{5pt}}ccc@{\hspace{5pt}}ccc@{\hspace{5pt}}ccccc@{\hspace{5pt}}c@{}}
\toprule
\textbf{Recurrent mixer} &
\multicolumn{2}{c}{\textbf{MMLU}} &
\multicolumn{3}{c}{\textbf{Code}} &
\multicolumn{3}{c}{\textbf{Math}} &
\multicolumn{5}{c}{\textbf{Commonsense understanding}} &
\textbf{Overall} \\
\cmidrule(lr){2-3}
\cmidrule(lr){4-6}
\cmidrule(lr){7-9}
\cmidrule(lr){10-14}
& \textbf{Pro} & \textbf{Str.}
& \textbf{HumanEval} & \textbf{MBPP} & \textbf{Avg.}
& \textbf{GSM8K} & \textbf{MATH-500} & \textbf{Avg.}
& \textbf{RACE} & \textbf{ARC-c} & \textbf{Hella.} & \textbf{Wino.} & \textbf{Avg.}
& \textbf{Avg.} \\
\midrule
\hspace{1.2mm}Mamba-2
& 37.41 & 61.27
& 42.71 & 56.32 & 49.52
& 71.87 & 60.80 & 66.34
& 75.50 & 77.56 & 77.78 & 70.56 & 75.35
& 60.14 \\
\hspace{1.2mm}KDA
& 37.16 & 60.30
& 42.83 & 58.72 & 50.78
& 74.68 & 64.40 & 69.54
& \textbf{78.47} & 78.16 & 77.94 & 69.14 & 75.93
& 61.25 \\
\hspace{1.2mm}Gated DeltaNet-2
& \textbf{38.00} & \textbf{61.73}
& \textbf{43.31} & \textbf{60.06} & \textbf{51.69}
& \textbf{79.30} & \textbf{65.20} & \textbf{72.25}
& 78.18 & \textbf{80.12} & \textbf{78.03} & \textbf{71.11} & \textbf{76.86}
& \textbf{62.67} \\
\bottomrule
\end{tabular}
\caption{
Accuracy of 31.6B-parameter hybrid latent-MoE models with 3.2B active parameters, trained on 300B tokens with full global-attention layers.
Only the recurrent mixer is varied; the attention layers, recurrent-state size, active parameter count, and training recipe are fixed.
Code Avg.\ averages HumanEval and MBPP, Math Avg.\ averages GSM8K and MATH-500, and the commonsense Avg.\ averages RACE, ARC-c, HellaSwag, and WinoGrande.
Overall Avg.\ equally averages the MMLU, code, math, and commonsense category scores, where the MMLU score is the mean of its two columns.
Best values are bolded.
}
\label{tab_scaling_hybrid}
\end{table*}

\paragraph{Scaling to a global-attention hybrid}
The experiments above fix the scale at 1.3B parameters and 100B tokens and pair each recurrent mixer with sliding-window attention. This leaves open whether the benefit of decoupled erase and write survives a larger model and training budget, and whether the finer memory edit still matters once a model contains full global-attention layers, which give direct access to the entire context and thus remove part of the pressure on the recurrent state. To answer both questions, we train Nemotron-3 Nano-style hybrid latent-MoE models~\citep{nvidia2025nemotron3,nvidia2025nemotron3nano} with 31.6B total and 3.2B active parameters on 300B tokens. Each model interleaves recurrent mixers and MoE layers with a small number of full global-attention layers. We keep the global-attention layers and their placement, the recurrent-state size, the active parameter count, and the training recipe fixed, and vary only the recurrent mixer among Mamba-2, KDA, and Gated DeltaNet-2. Small adjustments to the FFN intermediate size keep the three variants iso-parameter. Table~\ref{tab_scaling_hybrid} reports accuracy on knowledge, code, math, and commonsense benchmarks, which are described in Appendix~\ref{app_large_scale_setting}.

Gated DeltaNet-2 is best on nine of the ten benchmarks and on every category average. It improves the overall average by 1.42 points over KDA and by 2.53 points over Mamba-2. The gains are largest on math, where the category average improves by 2.71 and 5.91 points, respectively, and they extend to code and commonsense understanding. Two observations follow. First, the advantage of decoupling erase and write does not disappear with greater model capacity or a threefold larger token budget. The relevant quantity is not the parameter count alone but the number of competing associations that must share a fixed-size state, and larger models trained on more data can also learn more selective token-dependent erase and write policies. Second, global attention does not make the recurrent update redundant. Even when a few attention layers provide direct long-range access, replacing the recurrent mixer with Gated DeltaNet-2 yields consistent gains at matched state size, parameter budget, and training tokens. We therefore expect similar gains in larger global-attention hybrids, and potentially larger gains in recurrent-only or bounded-attention models at longer contexts, where interference within the fixed-size state is more severe. The latter remains a mechanistic expectation rather than a claim established by these experiments.

\paragraph{Gate structure and erase range ablations.}
\label{para_ablation}

Table~\ref{tab_ablation} evaluates two aspects of the Gated Delta Rule-2 update, the channel structure of the erase and write gates, and the range of the erase gate. For the channel-structure ablations, we average either gate over its channel axis and broadcast the scalar back at runtime, while keeping the original projections unchanged. Thus the parameter count stays fixed and only channel-wise gate variation is removed. Both scalarized variants trail full Gated DeltaNet-2, showing that both gates use their channel degrees of freedom. The asymmetry is clear. Keeping channel structure only in $\vb_t$ recovers most of the full model on language modeling and retrieval, whereas keeping it only in $\vw_t$ recovers less. This matches Eq.~\ref{eq_gdn2_recurrence}, where $\vb_t$ changes the key-side erase factor $\vk_t(\vb_t\odot\vk_t)^\top$, while $\vw_t$ reweights the written value. Finally, expanding the erase range from $[0,1]^{d_k}$ to $[0,2]^{d_k}$ gives no consistent gain at this scale.

\begin{table}[t]
\centering
\scriptsize
\addtolength{\tabcolsep}{-1.5pt}
\begin{tabular}{l|cc|c|ccc|c}
\toprule
\textbf{Variant} & \textbf{Wiki.} & \textbf{LMB.} & \textbf{Common.} &
\textbf{S-NIAH-2} & \textbf{S-NIAH-3} & \textbf{MK-NIAH-1} & \textbf{Recall} \\
 & ppl $\downarrow$ & ppl $\downarrow$ & avg $\uparrow$ &
@4K $\uparrow$ & @2K $\uparrow$ & @4K $\uparrow$ & avg $\uparrow$ \\
\midrule
\multicolumn{8}{l}{\emph{Channel structure}} \\
w-only, scalar $\vb_t$, channel $\vw_t$   & 16.55 & 11.62 & 52.45 & 90.6 & 71.4 & 30.6 & 28.92 \\
b-only, channel $\vb_t$, scalar $\vw_t$   & 16.12 & 11.50 & 52.79 & 92.1 & 84.6 & 35.2 & 29.51 \\
\midrule
\multicolumn{8}{l}{\emph{Erase range}} \\
Gated DeltaNet-2, $\vb_t \in [0,1]^{d_k}$     & \textbf{15.90} & \textbf{11.41} & \textbf{53.11} & 93.0 & \textbf{89.8} & \textbf{37.8} & \textbf{29.88} \\
\hspace{2mm} expanded $\vb_t \in [0,2]^{d_k}$ & 15.95 & 11.44 & 53.04 & \textbf{93.1} & 89.4 & 37.6 & 29.81 \\
\midrule
\multicolumn{8}{l}{\emph{Delta correction}} \\
Read/write gates, no delta correction & 16.28 & 11.60 & 52.58 & 85.4 & 72.1 & 30.9 & 29.08 \\
\bottomrule
\end{tabular}
\addtolength{\tabcolsep}{1.5pt}
\vspace{4pt}
\caption{
Gate structure, erase range, and delta-correction ablations in the recurrent-only setting.
The no-delta variant keeps both channel-wise gate projections but replaces the read-before-write correction with an additive memory, Eq.~\ref{eq_no_delta}.
}
\label{tab_ablation}
\end{table}

\paragraph{Removing the delta-rule correction.}
The ablations above all retain the delta rule, so they do not show whether the retrieval gains come from the read-before-write correction or from channel-wise gating alone. We therefore train a parameter-matched control that keeps both channel-wise gate projections but replaces the correction with an additive memory. Without an internal state read, $\vb_t$ becomes a channel-wise read gate on the query and $\vw_t$ remains a channel-wise write gate,
\begin{align}
    \bar{\rmS}_t=\rmD_t\rmS_{t-1},
    \qquad
    \rmS_t=\bar{\rmS}_t+\vk_t(\vw_t\odot\vv_t)^\top,
    \qquad
    \vo_t=\rmS_t^\top(\vb_t\odot\vq_t).
\label{eq_no_delta}
\end{align}
This control has the same recurrent-state size, parameter count, gate projections, and training recipe as Gated DeltaNet-2. The last block of Table~\ref{tab_ablation} shows that it remains competitive on language modeling, but degrades substantially when memory interference requires targeted replacement. Gated DeltaNet-2 improves S-NIAH-2, S-NIAH-3, and MK-NIAH-1 by 7.6, 17.7, and 6.9 points, respectively. This isolates the benefit of the delta correction from the additional vector gates: the retrieval gains of Gated DeltaNet-2 cannot be explained by channel-wise read and write gating alone.

\paragraph{Throughput comparison.}

Fig.~\ref{fig:throughput_hybrid} reports single H100 training throughput for the hybrid 1.3B models under a fixed token budget. Gated DeltaNet-2 preserves the near-flat scaling profile of recurrent mixers as sequence length grows, dropping only mildly from 38.0 to 36.1 Kt/s, while the 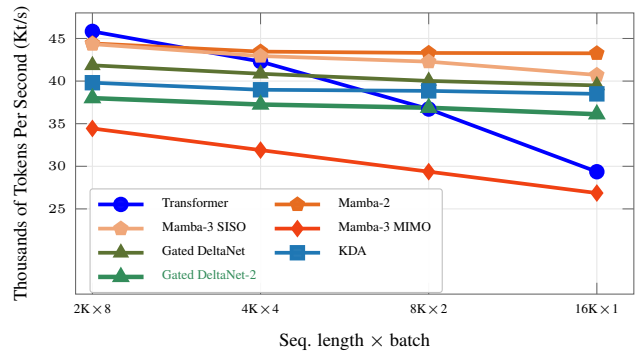
\begin{wrapfigure}[11]{r}{0.54\linewidth}
\vspace{-1.1\baselineskip}
\centering
\setlength{\intextsep}{4pt}
\setlength{\abovecaptionskip}{2pt}
\setlength{\belowcaptionskip}{-4pt}
\definecolor{gdngreen}{RGB}{92,114,50}
\definecolor{gdn2green}{RGB}{50,140,90}
\definecolor{mambaorange}{RGB}{230,108,32}
\definecolor{kdablue}{RGB}{31,119,180}
\begin{tikzpicture}
\begin{axis}[
    width=\linewidth,
    height=5.3cm,
    xlabel={Seq. length $\times$ batch},
    ylabel={Thousands of Tokens Per Second (Kt/s)},
    grid=major,
    major grid style={black!10},
    axis line style={black!60},
    tick style={black!60},
    xmin=0.9, xmax=4.22,
    ymin=15, ymax=48,
    xtick={1,2,3,4},
    xticklabels={2K$\times$8,4K$\times$4,8K$\times$2,16K$\times$1},
    ytick={25,30,35,40,45},
    tick label style={font=\tiny},
    label style={font=\scriptsize},
    every axis plot/.append style={line width=1.4pt, mark size=2.2pt},
    legend columns=2,
    legend cell align=left,
    legend style={
        at={(0.025,0.004)},
        anchor=south west,
        font=\fontsize{5.0}{5.6}\selectfont,
        draw=black!25,
        fill=white,
        fill opacity=0.96,
        text opacity=1,
        rounded corners=1pt,
        inner xsep=2.2pt,
        inner ysep=1.4pt,
        row sep=0.2pt,
        column sep=4.2pt
    },
    clip=true,
]

\addplot[blue, mark=*] coordinates {
    (1,45.83) (2,42.29) (3,36.72) (4,29.36)
};
\addlegendentry{Transformer}

\addplot[color=mambaorange, mark=pentagon*] coordinates {
    (1,44.42) (2,43.46) (3,43.30) (4,43.26)
};
\addlegendentry{Mamba-2}

\addplot[color=mambaorange!60, mark=pentagon*] coordinates {
    (1,44.34) (2,42.94) (3,42.28) (4,40.72)
};
\addlegendentry{Mamba-3 SISO}

\addplot[color=mambaorange!60!red, mark=diamond*] coordinates {
    (1,34.44) (2,31.90) (3,29.37) (4,26.86)
};
\addlegendentry{Mamba-3 MIMO}

\addplot[color=gdngreen, mark=triangle*] coordinates {
    (1,41.85) (2,40.86) (3,40.02) (4,39.49)
};
\addlegendentry{Gated DeltaNet}

\addplot[color=kdablue, mark=square*] coordinates {
    (1,39.81) (2,38.98) (3,38.85) (4,38.50)
};
\addlegendentry{KDA}

\addplot[color=gdn2green, mark=triangle*, very thick, line width=1.7pt] coordinates {
    (1,38.00) (2,37.25) (3,36.87) (4,36.11)
};
\addlegendentry{\textcolor{gdn2green}{Gated DeltaNet-2}}

\end{axis}
\end{tikzpicture}
\caption{Training throughput on a H100 GPU.}
\label{fig:throughput_hybrid}
\vspace{-1.4\baselineskip}
\end{wrapfigure}Transformer degrades sharply. Relative to KDA, the small gap reflects the added channel-wise erase and write gates. Thus Gated DeltaNet-2 retains practical training efficiency while paying a modest constant cost for finer memory control.

\section{Related Work}
\label{sec_related_work}

Efficient sequence models replace quadratic self-attention with recurrent or linear-time token mixers that maintain a fixed-size state. Early structured state-space and recurrent models used mostly data-independent transitions~\citep{s4,s5,Orvieto2023ResurrectingRN,sun2023retentive}, while Mamba and Mamba-2 introduced data-dependent selective dynamics and the SSD framework~\citep{gu_mamba_2023,pmlr-v235-dao24a}. Gated linear attention and related linear RNNs further improve memory control with learned decay gates~\citep{yang_gated_2023,qin2024hgrn2}. Delta-rule models take a complementary fast-weight view, where the recurrent state is updated by correcting the current read before writing the new value, improving associative memory over Hebbian-style accumulation~\citep{Gardner1988TheSO,Prados1989NeuralNC,irie2021going,yang2024parallelizing}. Gated DeltaNet adds adaptive forgetting to this update~\citep{yang2025gated}, and KDA strengthens it with channel-wise decay and an efficient chunkwise algorithm, but still uses a scalar $\beta_t$ to control both erasing and writing~\citep{team2025kimi}. Mamba-3 advances the SSM line instead, using exponential-trapezoidal discretization, complex-valued transitions implemented through data-dependent rotations, and a MIMO formulation for stronger modeling at efficient decoding latency~\citep{lahoti2026mamba3}. Our work is complementary to these directions. Gated DeltaNet-2 keeps the delta-rule fast-weight structure of GDN and KDA, but replaces the tied scalar update strength with a channel-wise erase gate $\vb_t$ and a channel-wise write gate $\vw_t$. This recovers KDA when both gates are tied to the same scalar, while allowing old content and new values to be controlled along different channel patterns.

\section{Conclusion}
\label{sec_conclusion}

We introduced Gated DeltaNet-2, a delta-rule recurrent attention layer that decouples the active memory edit into channel-wise erase and write decisions. The erase gate $\vb_t$ selects which key-side coordinates of the decayed state are read and removed, while the write gate $\vw_t$ selects which value-side coordinates are committed. This removes the scalar $\beta_t$ tie in Gated DeltaNet and KDA, recovers both as special cases, and preserves efficient chunkwise training through a WY form with gate-aware kernels. Under matched 1.3B training, Gated DeltaNet-2 improves the recurrent and hybrid frontier across language modeling, commonsense reasoning, synthetic retrieval, and real-world recall, while adding only a small constant throughput overhead. The gain persists in a 31.6B-parameter hybrid latent-MoE model with full global attention trained on 300B tokens, where Gated DeltaNet-2 leads KDA and Mamba-2 on every category average. Ablations show that both gates contribute, with the erase gate accounting for most of the gain.

\section*{Acknowledgements and Disclosure of Funding}
This work was conducted at and funded by NVIDIA; all authors are employees of NVIDIA. No third-party funding supported this work in the last 36 months, and the authors declare no financial relationships outside the submitted work that could be perceived as influencing it.

\bibliographystyle{unsrtnat}
\bibliography{ref}

@inproceedings{paperno_lambada_2016,
 address = {Berlin, Germany},
 author = {Paperno, Denis  and
Kruszewski, Germ{\'a}n  and
Lazaridou, Angeliki  and
Pham, Ngoc Quan  and
Bernardi, Raffaella  and
Pezzelle, Sandro  and
Baroni, Marco  and
Boleda, Gemma  and
Fern{\'a}ndez, Raquel},
 booktitle = {Proceedings of the 54th Annual Meeting of the Association for Computational Linguistics (Volume 1: Long Papers)},
 doi = {10.18653/v1/P16-1144},
 editor = {Erk, Katrin  and
Smith, Noah A.},
 pages = {1525--1534},
 publisher = {Association for Computational Linguistics},
 title = {The {LAMBADA} dataset: Word prediction requiring a broad discourse context},
 url = {https://aclanthology.org/P16-1144},
 year = {2016}
}

@inproceedings{sap2019social,
 address = {Hong Kong, China},
 author = {Sap, Maarten  and
Rashkin, Hannah  and
Chen, Derek  and
Le Bras, Ronan  and
Choi, Yejin},
 booktitle = {Proceedings of the 2019 Conference on Empirical Methods in Natural Language Processing and the 9th International Joint Conference on Natural Language Processing (EMNLP-IJCNLP)},
 doi = {10.18653/v1/D19-1454},
 editor = {Inui, Kentaro  and
Jiang, Jing  and
Ng, Vincent  and
Wan, Xiaojun},
 pages = {4463--4473},
 publisher = {Association for Computational Linguistics},
 title = {Social {IQ}a: Commonsense Reasoning about Social Interactions},
 url = {https://aclanthology.org/D19-1454},
 year = {2019}
}

@inproceedings{merity2016pointer,
 author = {Stephen Merity and
Caiming Xiong and
James Bradbury and
Richard Socher},
 bibsource = {dblp computer science bibliography, https://dblp.org},
 booktitle = {5th International Conference on Learning Representations, {ICLR} 2017,
Toulon, France, April 24-26, 2017, Conference Track Proceedings},
 publisher = {OpenReview.net},
 title = {Pointer Sentinel Mixture Models},
 url = {https://openreview.net/forum?id=Byj72udxe},
 year = {2017}
}

@inproceedings{rajpurkar_know_2018,
 address = {Melbourne, Australia},
 author = {Rajpurkar, Pranav  and
Jia, Robin  and
Liang, Percy},
 booktitle = {Proceedings of the 56th Annual Meeting of the Association for Computational Linguistics (Volume 2: Short Papers)},
 doi = {10.18653/v1/P18-2124},
 editor = {Gurevych, Iryna  and
Miyao, Yusuke},
 pages = {784--789},
 publisher = {Association for Computational Linguistics},
 title = {Know What You Don{'}t Know: Unanswerable Questions for {SQ}u{AD}},
 url = {https://aclanthology.org/P18-2124},
 year = {2018}
}

@inproceedings{lockard_openceres_2019,
 address = {Minneapolis, Minnesota},
 author = {Lockard, Colin  and
Shiralkar, Prashant  and
Dong, Xin Luna},
 booktitle = {Proceedings of the 2019 Conference of the North {A}merican Chapter of the Association for Computational Linguistics: Human Language Technologies, Volume 1 (Long and Short Papers)},
 doi = {10.18653/v1/N19-1309},
 editor = {Burstein, Jill  and
Doran, Christy  and
Solorio, Thamar},
 pages = {3047--3056},
 publisher = {Association for Computational Linguistics},
 title = {{O}pen{C}eres: {W}hen Open Information Extraction Meets the Semi-Structured Web},
 url = {https://aclanthology.org/N19-1309},
 year = {2019}
}

@misc{arora_language_2023,
 author = {Arora, Simran and Yang, Brandon and Eyuboglu, Sabri and Narayan, Avanika and Hojel, Andrew and Trummer, Immanuel and Ré, Christopher},
 journal = {ArXiv preprint},
 title = {Language {Models} {Enable} {Simple} {Systems} for {Generating} {Structured} {Views} of {Heterogeneous} {Data} {Lakes}},
 url = {https://arxiv.org/abs/2304.09433},
 volume = {abs/2304.09433},
 year = {2023}
}

@misc{de_griffin_2024,
 author = {De, Soham and Smith, Samuel L. and Fernando, Anushan and Botev, Aleksandar and Cristian-Muraru, George and Gu, Albert and Haroun, Ruba and Berrada, Leonard and Chen, Yutian and Srinivasan, Srivatsan and Desjardins, Guillaume and Doucet, Arnaud and Budden, David and Teh, Yee Whye and Pascanu, Razvan and De Freitas, Nando and Gulcehre, Caglar},
 journal = {ArXiv preprint},
 title = {Griffin: {Mixing} {Gated} {Linear} {Recurrences} with {Local} {Attention} for {Efficient} {Language} {Models}},
 url = {https://arxiv.org/abs/2402.19427},
 volume = {abs/2402.19427},
 year = {2024}
}

@inproceedings{DBLP:conf/icml/MerrillPS24,
 author = {William Merrill and
Jackson Petty and
Ashish Sabharwal},
 bibsource = {dblp computer science bibliography, https://dblp.org},
 booktitle = {Forty-first International Conference on Machine Learning, {ICML} 2024,
Vienna, Austria, July 21-27, 2024},
 publisher = {OpenReview.net},
 title = {The Illusion of State in State-Space Models},
 url = {https://openreview.net/forum?id=QZgo9JZpLq},
 year = {2024}
}

@inproceedings{ttt,
  title     = {Learning to (Learn at Test Time): RNNs with Expressive Hidden States},
  author    = {Sun, Yu and Li, Xinhao and Dalal, Karan and Xu, Jiarui and Vikram, Arjun and Zhang, Genghan and Dubois, Yann and Chen, Xinlei and Wang, Xiaolong and Koyejo, Sanmi and Hashimoto, Tatsunori and Guestrin, Carlos},
  booktitle = {Proceedings of the 42nd International Conference on Machine Learning},
  pages     = {57503--57522},
  year      = {2025},
  volume    = {267},
  series    = {Proceedings of Machine Learning Research},
  publisher = {PMLR},
  url       = {https://proceedings.mlr.press/v267/sun25h.html}
}

@article{Gardner1988TheSO,
 author = {E. Gardner},
 journal = {Journal of Physics A},
 pages = {257-270},
 title = {The space of interactions in neural network models},
 url = {https://api.semanticscholar.org/CorpusID:15378089},
 volume = {21},
 year = {1988}
}

@inproceedings{Irie2022TheDF,
 author = {Kazuki Irie and
R{\'{o}}bert Csord{\'{a}}s and
J{\"{u}}rgen Schmidhuber},
 bibsource = {dblp computer science bibliography, https://dblp.org},
 booktitle = {International Conference on Machine Learning, {ICML} 2022, 17-23 July
2022, Baltimore, Maryland, {USA}},
 editor = {Kamalika Chaudhuri and
Stefanie Jegelka and
Le Song and
Csaba Szepesv{\'{a}}ri and
Gang Niu and
Sivan Sabato},
 pages = {9639--9659},
 publisher = {{PMLR}},
 series = {Proceedings of Machine Learning Research},
 title = {The Dual Form of Neural Networks Revisited: Connecting Test Time Predictions
to Training Patterns via Spotlights of Attention},
 url = {https://proceedings.mlr.press/v162/irie22a.html},
 volume = {162},
 year = {2022}
}

@inproceedings{Orvieto2023ResurrectingRN,
 author = {Antonio Orvieto and
Samuel L. Smith and
Albert Gu and
Anushan Fernando and
{\c{C}}aglar G{\"{u}}l{\c{c}}ehre and
Razvan Pascanu and
Soham De},
 bibsource = {dblp computer science bibliography, https://dblp.org},
 booktitle = {International Conference on Machine Learning, {ICML} 2023, 23-29 July
2023, Honolulu, Hawaii, {USA}},
 editor = {Andreas Krause and
Emma Brunskill and
Kyunghyun Cho and
Barbara Engelhardt and
Sivan Sabato and
Jonathan Scarlett},
 pages = {26670--26698},
 publisher = {{PMLR}},
 series = {Proceedings of Machine Learning Research},
 title = {Resurrecting Recurrent Neural Networks for Long Sequences},
 url = {https://proceedings.mlr.press/v202/orvieto23a.html},
 volume = {202},
 year = {2023}
}

@inproceedings{akyurek_-context_2024,
  title     = {In-Context Language Learning: Architectures and Algorithms},
  author    = {Aky{\"u}rek, Ekin and Wang, Bailin and Kim, Yoon and Andreas, Jacob},
  booktitle = {Proceedings of the 41st International Conference on Machine Learning},
  pages     = {787--812},
  year      = {2024},
  volume    = {235},
  series    = {Proceedings of Machine Learning Research},
  publisher = {PMLR},
  url       = {https://proceedings.mlr.press/v235/akyurek24a.html}
}

@misc{peng_eagle_2024,
 author = {Peng, Bo and Goldstein, Daniel and Anthony, Quentin and Albalak, Alon and Alcaide, Eric and Biderman, Stella and Cheah, Eugene and Du, Xingjian and Ferdinan, Teddy and Hou, Haowen and Kazienko, Przemysław and GV, Kranthi Kiran and Kocoń, Jan and Koptyra, Bartłomiej and Krishna, Satyapriya and McClelland Jr., Ronald and Muennighoff, Niklas and Obeid, Fares and Saito, Atsushi and Song, Guangyu and Tu, Haoqin and Woźniak, Stanisław and Zhang, Ruichong and Zhao, Bingchen and Zhao, Qihang and Zhou, Peng and Zhu, Jian and Zhu, Rui-Jie},
 journal = {ArXiv preprint},
 title = {Eagle and {Finch}: {RWKV} with {Matrix}-{Valued} {States} and {Dynamic} {Recurrence}},
 url = {https://arxiv.org/abs/2404.05892},
 volume = {abs/2404.05892},
 year = {2024}
}

@inproceedings{tillet_triton_2019,
 author = {Tillet, Philippe and Kung, Hsiang-Tsung and Cox, David D.},
 booktitle = {Proceedings of the 3rd {ACM} {SIGPLAN} {International} {Workshop} on {Machine} {Learning} and {Programming} {Languages}},
 publisher = {ACM},
 title = {Triton: an intermediate language and compiler for tiled neural network computations},
 year = {2019}
}

@article{beck2024xlstm,
 author = {Beck, Maximilian and P{\"o}ppel, Korbinian and Spanring, Markus and Auer, Andreas and Prudnikova, Oleksandra and Kopp, Michael and Klambauer, G{\"u}nter and Brandstetter, Johannes and Hochreiter, Sepp},
 journal = {ArXiv preprint},
 title = {xLSTM: Extended Long Short-Term Memory},
 url = {https://arxiv.org/abs/2405.04517},
 volume = {abs/2405.04517},
 year = {2024}
}

@inproceedings{zoology,
  title     = {Zoology: Measuring and Improving Recall in Efficient Language Models},
  author    = {Arora, Simran and Eyuboglu, Sabri and Timalsina, Aman and Johnson, Isys and Poli, Michael and Zou, James and Rudra, Atri and R{\'e}, Christopher},
  booktitle = {The Twelfth International Conference on Learning Representations},
  year      = {2024},
  url       = {https://openreview.net/forum?id=LY3ukUANko}
}

@inproceedings{s5,
 author = {Jimmy T. H. Smith and
Andrew Warrington and
Scott W. Linderman},
 bibsource = {dblp computer science bibliography, https://dblp.org},
 booktitle = {The Eleventh International Conference on Learning Representations,
{ICLR} 2023, Kigali, Rwanda, May 1-5, 2023},
 publisher = {OpenReview.net},
 title = {Simplified State Space Layers for Sequence Modeling},
 url = {https://openreview.net/pdf?id=Ai8Hw3AXqks},
 year = {2023}
}

@inproceedings{s4,
 author = {Albert Gu and
Karan Goel and
Christopher R{\'{e}}},
 bibsource = {dblp computer science bibliography, https://dblp.org},
 booktitle = {The Tenth International Conference on Learning Representations, {ICLR}
2022, Virtual Event, April 25-29, 2022},
 publisher = {OpenReview.net},
 title = {Efficiently Modeling Long Sequences with Structured State Spaces},
 url = {https://openreview.net/forum?id=uYLFoz1vlAC},
 year = {2022}
}

@article{sun2023retentive,
 author = {Sun, Yutao and Dong, Li and Huang, Shaohan and Ma, Shuming and Xia, Yuqing and Xue, Jilong and Wang, Jianyong and Wei, Furu},
 journal = {ArXiv preprint},
 title = {Retentive network: A successor to transformer for large language models},
 url = {https://arxiv.org/abs/2307.08621},
 volume = {abs/2307.08621},
 year = {2023}
}

@inproceedings{linear-xmr-fastweight,
 author = {Imanol Schlag and
Kazuki Irie and
J{\"{u}}rgen Schmidhuber},
 bibsource = {dblp computer science bibliography, https://dblp.org},
 booktitle = {Proceedings of the 38th International Conference on Machine Learning,
{ICML} 2021, 18-24 July 2021, Virtual Event},
 editor = {Marina Meila and
Tong Zhang},
 pages = {9355--9366},
 publisher = {{PMLR}},
 series = {Proceedings of Machine Learning Research},
 title = {Linear Transformers Are Secretly Fast Weight Programmers},
 url = {http://proceedings.mlr.press/v139/schlag21a.html},
 volume = {139},
 year = {2021}
}

@inproceedings{irie2021going,
 author = {Kazuki Irie and
Imanol Schlag and
R{\'{o}}bert Csord{\'{a}}s and
J{\"{u}}rgen Schmidhuber},
 bibsource = {dblp computer science bibliography, https://dblp.org},
 booktitle = {Advances in Neural Information Processing Systems 34: Annual Conference
on Neural Information Processing Systems 2021, NeurIPS 2021, December
6-14, 2021, virtual},
 editor = {Marc'Aurelio Ranzato and
Alina Beygelzimer and
Yann N. Dauphin and
Percy Liang and
Jennifer Wortman Vaughan},
 pages = {7703--7717},
 title = {Going Beyond Linear Transformers with Recurrent Fast Weight Programmers},
 url = {https://proceedings.neurips.cc/paper/2021/hash/3f9e3767ef3b10a0de4c256d7ef9805d-Abstract.html},
 year = {2021}
}

@inproceedings{zellers2019hellaswag,
 address = {Florence, Italy},
 author = {Zellers, Rowan  and
Holtzman, Ari  and
Bisk, Yonatan  and
Farhadi, Ali  and
Choi, Yejin},
 booktitle = {Proceedings of the 57th Annual Meeting of the Association for Computational Linguistics},
 doi = {10.18653/v1/P19-1472},
 editor = {Korhonen, Anna  and
Traum, David  and
M{\`a}rquez, Llu{\'\i}s},
 pages = {4791--4800},
 publisher = {Association for Computational Linguistics},
 title = {{H}ella{S}wag: Can a Machine Really Finish Your Sentence?},
 url = {https://aclanthology.org/P19-1472},
 year = {2019}
}

@inproceedings{bisk2020piqa,
 author = {Yonatan Bisk and
Rowan Zellers and
Ronan LeBras and
Jianfeng Gao and
Yejin Choi},
 bibsource = {dblp computer science bibliography, https://dblp.org},
 booktitle = {The Thirty-Fourth {AAAI} Conference on Artificial Intelligence, {AAAI}
2020, The Thirty-Second Innovative Applications of Artificial Intelligence
Conference, {IAAI} 2020, The Tenth {AAAI} Symposium on Educational
Advances in Artificial Intelligence, {EAAI} 2020, New York, NY, USA,
February 7-12, 2020},
 pages = {7432--7439},
 publisher = {{AAAI} Press},
 title = {{PIQA:} Reasoning about Physical Commonsense in Natural Language},
 url = {https://aaai.org/ojs/index.php/AAAI/article/view/6239},
 year = {2020}
}

@article{arc-ce,
 author = {Clark, Peter and Cowhey, Isaac and Etzioni, Oren and Khot, Tushar and Sabharwal, Ashish and Schoenick, Carissa and Tafjord, Oyvind},
 journal = {ArXiv preprint},
 title = {Think you have solved question answering? try arc, the ai2 reasoning challenge},
 url = {https://arxiv.org/abs/1803.05457},
 volume = {abs/1803.05457},
 year = {2018}
}

@inproceedings{sakaguchi2021winogrande,
 author = {Keisuke Sakaguchi and
Ronan Le Bras and
Chandra Bhagavatula and
Yejin Choi},
 bibsource = {dblp computer science bibliography, https://dblp.org},
 booktitle = {The Thirty-Fourth {AAAI} Conference on Artificial Intelligence, {AAAI}
2020, The Thirty-Second Innovative Applications of Artificial Intelligence
Conference, {IAAI} 2020, The Tenth {AAAI} Symposium on Educational
Advances in Artificial Intelligence, {EAAI} 2020, New York, NY, USA,
February 7-12, 2020},
 pages = {8732--8740},
 publisher = {{AAAI} Press},
 title = {WinoGrande: An Adversarial Winograd Schema Challenge at Scale},
 url = {https://aaai.org/ojs/index.php/AAAI/article/view/6399},
 year = {2020}
}

@inproceedings{openbookqa,
 address = {Brussels, Belgium},
 author = {Mihaylov, Todor  and
Clark, Peter  and
Khot, Tushar  and
Sabharwal, Ashish},
 booktitle = {Proceedings of the 2018 Conference on Empirical Methods in Natural Language Processing},
 doi = {10.18653/v1/D18-1260},
 editor = {Riloff, Ellen  and
Chiang, David  and
Hockenmaier, Julia  and
Tsujii, Jun{'}ichi},
 pages = {2381--2391},
 publisher = {Association for Computational Linguistics},
 title = {Can a Suit of Armor Conduct Electricity? A New Dataset for Open Book Question Answering},
 url = {https://aclanthology.org/D18-1260},
 year = {2018}
}

@inproceedings{clark2019boolq,
 address = {Minneapolis, Minnesota},
 author = {Clark, Christopher  and
Lee, Kenton  and
Chang, Ming-Wei  and
Kwiatkowski, Tom  and
Collins, Michael  and
Toutanova, Kristina},
 booktitle = {Proceedings of the 2019 Conference of the North {A}merican Chapter of the Association for Computational Linguistics: Human Language Technologies, Volume 1 (Long and Short Papers)},
 doi = {10.18653/v1/N19-1300},
 editor = {Burstein, Jill  and
Doran, Christy  and
Solorio, Thamar},
 pages = {2924--2936},
 publisher = {Association for Computational Linguistics},
 title = {{B}ool{Q}: Exploring the Surprising Difficulty of Natural Yes/No Questions},
 url = {https://aclanthology.org/N19-1300},
 year = {2019}
}

@article{hsieh2024ruler,
 author = {Hsieh, Cheng-Ping and Sun, Simeng and Kriman, Samuel and Acharya, Shantanu and Rekesh, Dima and Jia, Fei and Zhang, Yang and Ginsburg, Boris},
 journal = {ArXiv preprint},
 title = {RULER: What's the Real Context Size of Your Long-Context Language Models?},
 url = {https://arxiv.org/abs/2404.06654},
 volume = {abs/2404.06654},
 year = {2024}
}

@article{Lieber2024JambaAH,
 author = {Lieber, Opher and Lenz, Barak and Bata, Hofit and Cohen, Gal and Osin, Jhonathan and Dalmedigos, Itay and Safahi, Erez and Meirom, Shaked and Belinkov, Yonatan and Shalev-Shwartz, Shai and others},
 journal = {ArXiv preprint},
 title = {Jamba: A hybrid transformer-mamba language model},
 url = {https://arxiv.org/abs/2403.19887},
 volume = {abs/2403.19887},
 year = {2024}
}

@article{Prados1989NeuralNC,
 author = {Prados, DL and Kak, SC},
 journal = {Electronics Letters},
 number = {25},
 pages = {197--199},
 title = {Neural network capacity using delta rule},
 volume = {3},
 year = {1989}
}

@inproceedings{Grazzi2024UnlockingSI,
  title     = {Unlocking State-Tracking in Linear RNNs Through Negative Eigenvalues},
  author    = {Grazzi, Riccardo and Siems, Julien and Zela, Arber and Franke, J{\"o}rg K. H. and Hutter, Frank and Pontil, Massimiliano},
  booktitle = {The Thirteenth International Conference on Learning Representations},
  year      = {2025},
  url       = {https://openreview.net/forum?id=UvTo3tVBk2}
}

@article{Joffrain2006AccumulatingHT,
 author = {Thierry Joffrain and Tze Meng Low and Enrique S. Quintana-Ort{\'i} and Robert A. van de Geijn and Field G. Van Zee},
 journal = {ACM Trans. Math. Softw.},
 pages = {169-179},
 title = {Accumulating Householder transformations, revisited},
 url = {https://api.semanticscholar.org/CorpusID:15723171},
 volume = {32},
 year = {2006}
}

@inproceedings{hua_transformer_2022,
 author = {Weizhe Hua and
Zihang Dai and
Hanxiao Liu and
Quoc V. Le},
 bibsource = {dblp computer science bibliography, https://dblp.org},
 booktitle = {International Conference on Machine Learning, {ICML} 2022, 17-23 July
2022, Baltimore, Maryland, {USA}},
 editor = {Kamalika Chaudhuri and
Stefanie Jegelka and
Le Song and
Csaba Szepesv{\'{a}}ri and
Gang Niu and
Sivan Sabato},
 pages = {9099--9117},
 publisher = {{PMLR}},
 series = {Proceedings of Machine Learning Research},
 title = {Transformer Quality in Linear Time},
 url = {https://proceedings.mlr.press/v162/hua22a.html},
 volume = {162},
 year = {2022}
}

@inproceedings{peng_rwkv_2023,
 address = {Singapore},
 author = {Peng, Bo  and
Alcaide, Eric  and
Anthony, Quentin  and
Albalak, Alon  and
Arcadinho, Samuel  and
Biderman, Stella  and
Cao, Huanqi  and
Cheng, Xin  and
Chung, Michael  and
Derczynski, Leon  and
Du, Xingjian  and
Grella, Matteo  and
Gv, Kranthi  and
He, Xuzheng  and
Hou, Haowen  and
Kazienko, Przemyslaw  and
Kocon, Jan  and
Kong, Jiaming  and
Koptyra, Bart{\l}omiej  and
Lau, Hayden  and
Lin, Jiaju  and
Mantri, Krishna Sri Ipsit  and
Mom, Ferdinand  and
Saito, Atsushi  and
Song, Guangyu  and
Tang, Xiangru  and
Wind, Johan  and
Wo{\'z}niak, Stanis{\l}aw  and
Zhang, Zhenyuan  and
Zhou, Qinghua  and
Zhu, Jian  and
Zhu, Rui-Jie},
 booktitle = {Findings of the Association for Computational Linguistics: EMNLP 2023},
 doi = {10.18653/v1/2023.findings-emnlp.936},
 editor = {Bouamor, Houda  and
Pino, Juan  and
Bali, Kalika},
 pages = {14048--14077},
 publisher = {Association for Computational Linguistics},
 title = {{RWKV}: Reinventing {RNN}s for the Transformer Era},
 url = {https://aclanthology.org/2023.findings-emnlp.936},
 year = {2023}
}

@inproceedings{arora-2024-jrt,
  title={Just read twice: closing the recall gap for recurrent language models},
  author={Arora, Simran and Timalsina, Aman and Singhal, Aaryan and Spector, Benjamin and Eyuboglu, Sabri and Zhao, Xinyi and Rao, Ashish and Rudra, Atri and R{\'e}, Christopher},
  booktitle={Proceedings of the 2nd Efficient Systems for Foundation Models Workshop at the International Conference on Machine Learning (ICML)},
  series={Proceedings of Machine Learning Research},
  volume={235},
  year={2024}
}

@inproceedings{katharopoulos_transformers_2020,
 author = {Angelos Katharopoulos and
Apoorv Vyas and
Nikolaos Pappas and
Fran{\c{c}}ois Fleuret},
 bibsource = {dblp computer science bibliography, https://dblp.org},
 booktitle = {Proceedings of the 37th International Conference on Machine Learning,
{ICML} 2020, 13-18 July 2020, Virtual Event},
 pages = {5156--5165},
 publisher = {{PMLR}},
 series = {Proceedings of Machine Learning Research},
 title = {Transformers are RNNs: Fast Autoregressive Transformers with Linear
Attention},
 url = {http://proceedings.mlr.press/v119/katharopoulos20a.html},
 volume = {119},
 year = {2020}
}

@article{hochreiter_long_1997,
 author = {Hochreiter, Sepp and Schmidhuber, Jürgen},
 journal = {Neural Computation},
 number = {8},
 pages = {1735--1780},
 title = {Long {Short}-{Term} {Memory}},
 volume = {9},
 year = {1997}
}

@inproceedings{gu_mamba_2023,
 author = {Gu, Albert and Dao, Tri},
 title = {Mamba: {Linear}-{Time} {Sequence} {Modeling} with {Selective} {State} {Spaces}},
 year = {2023}
}

@inproceedings{arora_simple_2024,
  title     = {Simple Linear Attention Language Models Balance the Recall-Throughput Tradeoff},
  author    = {Arora, Simran and Eyuboglu, Sabri and Zhang, Michael and Timalsina, Aman and Alberti, Silas and Zou, James and Rudra, Atri and R{\'e}, Christopher},
  booktitle = {Proceedings of the 41st International Conference on Machine Learning},
  pages     = {1763--1840},
  year      = {2024},
  volume    = {235},
  series    = {Proceedings of Machine Learning Research},
  publisher = {PMLR},
  url       = {https://proceedings.mlr.press/v235/arora24a.html}
}

@article{penedo2024fineweb,
 author = {Penedo, Guilherme and Kydl{\'\i}{\v{c}}ek, Hynek and Lozhkov, Anton and Mitchell, Margaret and Raffel, Colin and Von Werra, Leandro and Wolf, Thomas and others},
 journal = {ArXiv preprint},
 title = {The fineweb datasets: Decanting the web for the finest text data at scale},
 url = {https://arxiv.org/abs/2406.17557},
 volume = {abs/2406.17557},
 year = {2024}
}

@inproceedings{longhorn,
  title={Longhorn: State space models are amortized online learners},
  author={Liu, Bo and Wang, Rui and Wu, Lemeng and Feng, Yihao and Stone, Peter and Liu, Qiang},
  booktitle={International Conference on Learning Representations},
  volume={2025},
  pages={95419--95434},
  year={2025}
}

@inproceedings{yang2024parallelizing,
  title     = {Parallelizing Linear Transformers with the Delta Rule over Sequence Length},
  author    = {Yang, Songlin and Wang, Bailin and Zhang, Yu and Shen, Yikang and Kim, Yoon},
  booktitle = {Advances in Neural Information Processing Systems 37},
  year      = {2024},
  pages     = {115491--115522},
  url       = {https://proceedings.neurips.cc/paper_files/paper/2024/hash/d13a3eae72366e61dfdc7eea82eeb685-Abstract-Conference.html}
}

@inproceedings{ren2024samba,
  title     = {Samba: Simple Hybrid State Space Models for Efficient Unlimited Context Language Modeling},
  author    = {Ren, Liliang and Liu, Yang and Lu, Yadong and Shen, Yelong and Liang, Chen and Chen, Weizhu},
  booktitle = {The Thirteenth International Conference on Learning Representations},
  year      = {2025},
  url       = {https://openreview.net/forum?id=bIlnpVM4bc}
}

@article{qin2024hgrn2,
 author = {Qin, Zhen and Yang, Songlin and Sun, Weixuan and Shen, Xuyang and Li, Dong and Sun, Weigao and Zhong, Yiran},
 journal = {ArXiv preprint},
 title = {Hgrn2: Gated linear rnns with state expansion},
 url = {https://arxiv.org/abs/2404.07904},
 volume = {abs/2404.07904},
 year = {2024}
}

@inproceedings{yang_gated_2023,
 author = {Yang, Songlin and Wang, Bailin and Shen, Yikang and Panda, Rameswar and Kim, Yoon},
 booktitle = {Proceedings of the 41st International Conference on Machine Learning},
 pages = {56501--56523},
 publisher = {PMLR},
 series = {Proceedings of Machine Learning Research},
 title = {Gated Linear Attention Transformers with Hardware-Efficient Training},
 url = {https://proceedings.mlr.press/v235/yang24ab.html},
 volume = {235},
 year = {2024}
}

@inproceedings{pmlr-v235-dao24a,
 author = {Dao, Tri and Gu, Albert},
 booktitle = {Proceedings of the 41st International Conference on Machine Learning},
 pages = {10041--10071},
 publisher = {PMLR},
 series = {Proceedings of Machine Learning Research},
 title = {Transformers are {SSM}s: Generalized Models and Efficient Algorithms Through Structured State Space Duality},
 url = {https://proceedings.mlr.press/v235/dao24a.html},
 volume = {235},
 year = {2024}
}

@inproceedings{wen_rnns_2024,
  title     = {RNNs are not Transformers (Yet): The Key Bottleneck on In-context Retrieval},
  author    = {Wen, Kaiyue and Dang, Xingyu and Lyu, Kaifeng},
  booktitle = {The Thirteenth International Conference on Learning Representations},
  year      = {2025},
  url       = {https://openreview.net/forum?id=h3wbI8Uk1Z}
}

@inproceedings{bischof_wy_1985,
 author = {Bischof, Christian H. and Loan, Charles Van},
 booktitle = {{SIAM} {Conference} on {Parallel} {Processing} for {Scientific} {Computing}},
 title = {The {WY} representation for products of householder matrices},
 url = {https://api.semanticscholar.org/CorpusID:36094006},
 year = {1985}
}

@inproceedings{widrow_adaptive_1988,
 author = {Widrow, Bernard and Hoff, Marcian E and others},
 booktitle = {IRE WESCON convention record},
 number = {1},
 organization = {New York},
 pages = {96--104},
 title = {Adaptive switching circuits},
 volume = {4},
 year = {1960}
}

@article{47761,
 address = {Cambridge, MA},
 author = {Kwiatkowski, Tom  and
Palomaki, Jennimaria  and
Redfield, Olivia  and
Collins, Michael  and
Parikh, Ankur  and
Alberti, Chris  and
Epstein, Danielle  and
Polosukhin, Illia  and
Devlin, Jacob  and
Lee, Kenton  and
Toutanova, Kristina  and
Jones, Llion  and
Kelcey, Matthew  and
Chang, Ming-Wei  and
Dai, Andrew M.  and
Uszkoreit, Jakob  and
Le, Quoc  and
Petrov, Slav},
 doi = {10.1162/tacl_a_00276},
 editor = {Lee, Lillian  and
Johnson, Mark  and
Roark, Brian  and
Nenkova, Ani},
 journal = {Transactions of the Association for Computational Linguistics},
 pages = {452--466},
 publisher = {MIT Press},
 title = {Natural Questions: A Benchmark for Question Answering Research},
 url = {https://aclanthology.org/Q19-1026},
 volume = {7},
 year = {2019}
}

@inproceedings{jelassi_repeat_2024,
  title     = {Repeat After Me: Transformers are Better than State Space Models at Copying},
  author    = {Jelassi, Samy and Brandfonbrener, David and Kakade, Sham M. and Malach, Eran},
  booktitle = {Proceedings of the 41st International Conference on Machine Learning},
  pages     = {21502--21521},
  year      = {2024},
  volume    = {235},
  series    = {Proceedings of Machine Learning Research},
  publisher = {PMLR},
  url       = {https://proceedings.mlr.press/v235/jelassi24a.html}
}

@inproceedings{dua2019drop,
 address = {Minneapolis, Minnesota},
 author = {Dua, Dheeru  and
Wang, Yizhong  and
Dasigi, Pradeep  and
Stanovsky, Gabriel  and
Singh, Sameer  and
Gardner, Matt},
 booktitle = {Proceedings of the 2019 Conference of the North {A}merican Chapter of the Association for Computational Linguistics: Human Language Technologies, Volume 1 (Long and Short Papers)},
 doi = {10.18653/v1/N19-1246},
 editor = {Burstein, Jill  and
Doran, Christy  and
Solorio, Thamar},
 pages = {2368--2378},
 publisher = {Association for Computational Linguistics},
 title = {{DROP}: A Reading Comprehension Benchmark Requiring Discrete Reasoning Over Paragraphs},
 url = {https://aclanthology.org/N19-1246},
 year = {2019}
}

@inproceedings{JoshiTriviaQA2017,
 address = {Vancouver, Canada},
 author = {Joshi, Mandar  and
Choi, Eunsol  and
Weld, Daniel  and
Zettlemoyer, Luke},
 booktitle = {Proceedings of the 55th Annual Meeting of the Association for Computational Linguistics (Volume 1: Long Papers)},
 doi = {10.18653/v1/P17-1147},
 editor = {Barzilay, Regina  and
Kan, Min-Yen},
 pages = {1601--1611},
 publisher = {Association for Computational Linguistics},
 title = {{T}rivia{QA}: A Large Scale Distantly Supervised Challenge Dataset for Reading Comprehension},
 url = {https://aclanthology.org/P17-1147},
 year = {2017}
}

@inproceedings{lahoti2026mamba3,
  title     = {Mamba-3: Improved Sequence Modeling using State Space Principles},
  author    = {Lahoti, Aakash and Li, Kevin Y. and Chen, Berlin and Wang, Caitlin and Bick, Aviv and Kolter, J. Zico and Dao, Tri and Gu, Albert},
  booktitle = {The Fourteenth International Conference on Learning Representations},
  year      = {2026},
  url       = {https://openreview.net/forum?id=HwCvaJOiCj}
}

@inproceedings{yang2025gated,
  title     = {Gated Delta Networks: Improving Mamba2 with Delta Rule},
  author    = {Yang, Songlin and Kautz, Jan and Hatamizadeh, Ali},
  booktitle = {The Thirteenth International Conference on Learning Representations},
  year      = {2025},
  url       = {https://openreview.net/forum?id=r8H7xhYPwz}
}

@inproceedings{siems2025deltaproduct,
  title         = {DeltaProduct: Improving State-Tracking in Linear RNNs via Householder Products},
  author        = {Siems, Julien and Carstensen, Timur and Zela, Arber and Hutter, Frank and Pontil, Massimiliano and Grazzi, Riccardo},
  booktitle     = {Advances in Neural Information Processing Systems},
  year          = {2025},
  url           = {https://openreview.net/forum?id=SoRiaijTGr},
  eprint        = {2502.10297},
  archivePrefix = {arXiv}
}

@article{team2025kimi,
  title={Kimi linear: An expressive, efficient attention architecture},
  author={Team, Kimi and Zhang, Yu and Lin, Zongyu and Yao, Xingcheng and Hu, Jiaxi and Meng, Fanqing and Liu, Chengyin and Men, Xin and Yang, Songlin and Li, Zhiyuan and others},
  journal={arXiv preprint arXiv:2510.26692},
  year={2025}
}

@article{peng2025rwkv7,
  title   = {{RWKV-7} ``{Goose}'' with Expressive Dynamic State Evolution},
  author  = {Peng, Bo and Zhang, Ruichong and Goldstein, Daniel and Alcaide, Eric and Du, Xingjian and Hou, Haowen and Lin, Jiaju and Liu, Jiaxing and Lu, Janna and Merrill, William and Song, Guangyu and Tan, Kaifeng and Utpala, Saiteja and Wilce, Nathan and Wind, Johan S. and Wu, Tianyi and Wuttke, Daniel and Zhou-Zheng, Christian},
  journal = {arXiv preprint arXiv:2503.14456},
  year    = {2025}
}

@article{behrouz2025miras,
  title   = {It's All Connected: A Journey Through Test-Time Memorization, Attentional Bias, Retention, and Online Optimization},
  author  = {Behrouz, Ali and Razaviyayn, Meisam and Zhong, Peilin and Mirrokni, Vahab},
  journal = {arXiv preprint arXiv:2504.13173},
  year    = {2025}
}

@article{sun2026fg2gdn,
  title   = {{FG}$^2$-{GDN}: Enhancing Long-Context Gated Delta Networks with Doubly Fine-Grained Control},
  author  = {Sun, Pingwei and Hu, Yuxuan and Tan, Jianchao and Wang, Xue and Zhang, Jiaqi and Lu, Yifan and Sun, Yerui and Xie, Yuchen and Cai, Xunliang},
  journal = {arXiv preprint arXiv:2604.19021},
  year    = {2026}
}

@article{nvidia2025nemotron3,
  title   = {{NVIDIA} {Nemotron} 3: Efficient and Open Intelligence},
  author  = {{NVIDIA}},
  journal = {arXiv preprint arXiv:2512.20856},
  year    = {2025}
}

@article{nvidia2025nemotron3nano,
  title   = {{Nemotron} 3 {Nano}: Open, Efficient Mixture-of-Experts Hybrid {Mamba}-{Transformer} Model for Agentic Reasoning},
  author  = {{NVIDIA}},
  journal = {arXiv preprint arXiv:2512.20848},
  year    = {2025}
}

@inproceedings{hendrycks2021mmlu,
  title     = {Measuring Massive Multitask Language Understanding},
  author    = {Hendrycks, Dan and Burns, Collin and Basart, Steven and Zou, Andy and Mazeika, Mantas and Song, Dawn and Steinhardt, Jacob},
  booktitle = {International Conference on Learning Representations},
  year      = {2021}
}

@inproceedings{wang2024mmlupro,
  title     = {{MMLU-Pro}: A More Robust and Challenging Multi-Task Language Understanding Benchmark},
  author    = {Wang, Yubo and Ma, Xueguang and Zhang, Ge and Ni, Yuansheng and Chandra, Abhranil and Guo, Shiguang and Ren, Weiming and Arulraj, Aaran and He, Xuan and Jiang, Ziyan and Li, Tianle and Ku, Max and Wang, Kai and Zhuang, Alex and Fan, Rongqi and Yue, Xiang and Chen, Wenhu},
  booktitle = {Advances in Neural Information Processing Systems},
  year      = {2024}
}

@article{chen2021codex,
  title   = {Evaluating Large Language Models Trained on Code},
  author  = {Chen, Mark and Tworek, Jerry and Jun, Heewoo and Yuan, Qiming and Pinto, Henrique Ponde de Oliveira and Kaplan, Jared and Edwards, Harri and Burda, Yuri and Joseph, Nicholas and Brockman, Greg and others},
  journal = {arXiv preprint arXiv:2107.03374},
  year    = {2021}
}

@article{austin2021mbpp,
  title   = {Program Synthesis with Large Language Models},
  author  = {Austin, Jacob and Odena, Augustus and Nye, Maxwell and Bosma, Maarten and Michalewski, Henryk and Dohan, David and Jiang, Ellen and Cai, Carrie and Terry, Michael and Le, Quoc and Sutton, Charles},
  journal = {arXiv preprint arXiv:2108.07732},
  year    = {2021}
}

@article{cobbe2021gsm8k,
  title   = {Training Verifiers to Solve Math Word Problems},
  author  = {Cobbe, Karl and Kosaraju, Vineet and Bavarian, Mohammad and Chen, Mark and Jun, Heewoo and Kaiser, Lukasz and Plappert, Matthias and Tworek, Jerry and Hilton, Jacob and Nakano, Reiichiro and Hesse, Christopher and Schulman, John},
  journal = {arXiv preprint arXiv:2110.14168},
  year    = {2021}
}

@inproceedings{hendrycks2021math,
  title     = {Measuring Mathematical Problem Solving With the {MATH} Dataset},
  author    = {Hendrycks, Dan and Burns, Collin and Kadavath, Saurav and Arora, Akul and Basart, Steven and Tang, Eric and Song, Dawn and Steinhardt, Jacob},
  booktitle = {Advances in Neural Information Processing Systems, Datasets and Benchmarks Track},
  year      = {2021}
}

@inproceedings{lightman2024verify,
  title     = {Let's Verify Step by Step},
  author    = {Lightman, Hunter and Kosaraju, Vineet and Burda, Yuri and Edwards, Harrison and Baker, Bowen and Lee, Teddy and Leike, Jan and Schulman, John and Sutskever, Ilya and Cobbe, Karl},
  booktitle = {International Conference on Learning Representations},
  year      = {2024}
}

@inproceedings{lai2017race,
  title     = {{RACE}: Large-scale {ReAding} Comprehension Dataset From Examinations},
  author    = {Lai, Guokun and Xie, Qizhe and Liu, Hanxiao and Yang, Yiming and Hovy, Eduard},
  booktitle = {Proceedings of the 2017 Conference on Empirical Methods in Natural Language Processing},
  pages     = {785--794},
  year      = {2017}
}

\newpage
\appendix
\renewcommand{\thesection}{\Alph{section}}
\renewcommand\thefigure{S.\arabic{figure}}
\setcounter{figure}{0}
\renewcommand\thetable{S.\arabic{table}}
\setcounter{table}{0}

{\noindent\Large\bfseries Appendix\par}
\vspace{-0.6em}

\section{Chunkwise derivation for Gated DeltaNet-2}
\label{app_wy}

This appendix gives the exact chunkwise form used in Section~\ref{subsec_chunkwise_training}. We work inside one chunk of length $C$ and write $\rmS_0$ for the state at the start of the chunk. All vectors are for a single head. In particular, $\vk_r,\ve_r,\vq_r,\vb_r,\valpha_r,\boldsymbol{\gamma}_r\in\mathbb{R}^{d_k}$, $\vv_r,\vz_r,\vo_r,\vw_r\in\mathbb{R}^{d_v}$, and $\rmS_r\in\mathbb{R}^{d_k\times d_v}$. The chunk index is suppressed.

The Gated DeltaNet-2 recurrence is
\begin{align}
    \rmS_r
    =
    \bigl(\rmI - \vk_r\ve_r^\top\bigr)
    \operatorname{Diag}(\valpha_r)\rmS_{r-1}
    + \vk_r\vz_r^\top,
    \qquad
    \ve_r = \vb_r \odot \vk_r,
    \qquad
    \vz_r = \vw_r \odot \vv_r .
\label{eq_app_gdn2_step}
\end{align}
Let
\begin{align}
    \boldsymbol{G}_r = \sum_{i=1}^{r}\boldsymbol{g}_i,
    \qquad
    \boldsymbol{\gamma}_r = \exp(\boldsymbol{G}_r),
    \qquad
    \boldsymbol{\gamma}_0 = \mathbf{1}_{d_k},
    \qquad
    \valpha_r = \exp(\boldsymbol{g}_r).
\label{eq_app_gamma}
\end{align}
All exponentials, products, and ratios involving $\boldsymbol{\gamma}$ are elementwise over the key channel axis.

\subsection{Decay-normalized recurrence}
\label{app_decay_normalized}

Define a normalized state $\widehat{\rmS}_r$ by
\begin{align}
    \rmS_r = \operatorname{Diag}(\boldsymbol{\gamma}_r)\widehat{\rmS}_r .
\label{eq_app_normalized_state}
\end{align}
Because $\boldsymbol{\gamma}_0=\mathbf{1}_{d_k}$, the normalized initial state is also $\widehat{\rmS}_0=\rmS_0$. Substituting Eq.~\ref{eq_app_normalized_state} into Eq.~\ref{eq_app_gdn2_step} and using $\boldsymbol{\gamma}_r=\valpha_r\odot\boldsymbol{\gamma}_{r-1}$ gives
\begin{align}
    \widehat{\rmS}_r
    =
    \bigl(\rmI - \bar{\vk}_r\bar{\ve}_r^\top\bigr)
    \widehat{\rmS}_{r-1}
    + \bar{\vk}_r\vz_r^\top,
    \qquad
    \bar{\vk}_r = \boldsymbol{\gamma}_r^{-1}\odot\vk_r,
    \qquad
    \bar{\ve}_r = \boldsymbol{\gamma}_r\odot\ve_r .
\label{eq_app_normalized_recurrence}
\end{align}
The channel-wise decay has disappeared from the recurrence. It is now carried by the left and right factors of each rank-one edit.

Let $\rmK$, $\rmV$, $\rmB$, and $\rmW$ contain rows $\vk_r^\top$, $\vv_r^\top$, $\vb_r^\top$, and $\vw_r^\top$, respectively. For compact matrix notation, let $\boldsymbol{\gamma}$ contain rows $\boldsymbol{\gamma}_r^\top$. Let $\bar{\rmK}$, $\bar{\rmE}$, and $\rmZ$ contain rows $\bar{\vk}_r^\top$, $\bar{\ve}_r^\top$, and $\vz_r^\top$. Equivalently,
\begin{align}
    \bar{\rmK}
    =
    \boldsymbol{\gamma}^{-1}\odot\rmK,
    \qquad
    \bar{\rmE}
    =
    \boldsymbol{\gamma}\odot(\rmB\odot\rmK),
    \qquad
    \rmZ
    =
    \rmW\odot\rmV .
\label{eq_app_gate_matrices}
\end{align}
Define
\begin{align}
    \rmT = \operatorname{tril}(\bar{\rmE}\bar{\rmK}^\top, -1),
    \qquad
    \rmA = (\rmI + \rmT)^{-1},
    \qquad
    \rmY = \rmA\bar{\rmE},
    \qquad
    \rmU = \rmA\rmZ .
\label{eq_app_wy_definitions}
\end{align}
Since $\rmT$ is strictly lower triangular, $\rmA$ is lower triangular with unit diagonal and is obtained by forward substitution.

\subsection{Compact state formula}
\label{app_state_formula}

Define
\begin{align}
    \rmR = \rmU - \rmY\rmS_0 .
\label{eq_app_residual_matrix}
\end{align}
Let row $r$ of $\rmR$ be $\boldsymbol{\rho}_r^\top$, where $\boldsymbol{\rho}_r\in\mathbb{R}^{d_v}$. Then the normalized state after any prefix of the chunk is
\begin{align}
    \widehat{\rmS}_r
    =
    \rmS_0
    +
    \bar{\rmK}_{\le r}^{\top}\rmR_{\le r},
\label{eq_app_prefix_state}
\end{align}
where $\bar{\rmK}_{\le r}$ and $\rmR_{\le r}$ denote the first $r$ rows.

To prove Eq.~\ref{eq_app_prefix_state}, write the rank-one increment at step $r$ as $\bar{\vk}_r\boldsymbol{\rho}_r^\top$. From Eq.~\ref{eq_app_normalized_recurrence}, the residual row is
\begin{align}
    \boldsymbol{\rho}_r^\top
    =
    \vz_r^\top
    -
    \bar{\ve}_r^\top\widehat{\rmS}_{r-1}.
\label{eq_app_residual_row}
\end{align}
Equivalently, in column-vector form, $\boldsymbol{\rho}_r=\vz_r-\widehat{\rmS}_{r-1}^{\top}\bar{\ve}_r$. Using the induction hypothesis $\widehat{\rmS}_{r-1}=\rmS_0+\sum_{s<r}\bar{\vk}_s\boldsymbol{\rho}_s^\top$ gives
\begin{align}
    \boldsymbol{\rho}_r^\top
    =
    \vz_r^\top
    -
    \bar{\ve}_r^\top\rmS_0
    -
    \sum_{s<r}
    \bar{\ve}_r^\top\bar{\vk}_s\,\boldsymbol{\rho}_s^\top .
\label{eq_app_residual_row_expanded}
\end{align}
Since $\rmT_{rs}=\bar{\ve}_r^\top\bar{\vk}_s$ for $s<r$ and $\rmT_{rs}=0$ otherwise, stacking these residual rows over the chunk yields
\begin{align}
    (\rmI+\rmT)\rmR
    =
    \rmZ-\bar{\rmE}\rmS_0 .
\label{eq_app_residual_linear_system}
\end{align}
Multiplying by $\rmA$ gives $\rmR=\rmA\rmZ-\rmA\bar{\rmE}\rmS_0=\rmU-\rmY\rmS_0$, which is Eq.~\ref{eq_app_residual_matrix}. Substituting the resulting increments into the normalized recurrence proves Eq.~\ref{eq_app_prefix_state}.

Multiplying Eq.~\ref{eq_app_prefix_state} by $\operatorname{Diag}(\boldsymbol{\gamma}_C)$ gives the end-of-chunk state
\begin{align}
    \rmS_C
    =
    \operatorname{Diag}(\boldsymbol{\gamma}_C)\rmS_0
    +
    \rmK_{\mathrm{tail}}^\top
    \bigl(\rmU-\rmY\rmS_0\bigr),
\label{eq_app_chunk_state}
\end{align}
where row $r$ of $\rmK_{\mathrm{tail}}$ is
\begin{align}
    (\rmK_{\mathrm{tail}})_{r,:}
    =
    \bigl((\boldsymbol{\gamma}_C/\boldsymbol{\gamma}_r)
    \odot \vk_r\bigr)^\top .
\label{eq_app_ktail}
\end{align}
This is Eq.~\ref{eq_gdn2_chunk_state} in the main text.

\subsection{Compact output formula}
\label{app_output_formula}

The output at token $r$ is the column vector $\vo_r=\rmS_r^\top\vq_r$. It is convenient to write the corresponding row vector. Using Eq.~\ref{eq_app_prefix_state},
\begin{align}
    \vo_r^\top
    =
    (\boldsymbol{\gamma}_r\odot\vq_r)^\top\rmS_0
    +
    \sum_{s\le r}
    \bigl[
    \vq_r^\top
    \operatorname{Diag}(\boldsymbol{\gamma}_r/\boldsymbol{\gamma}_s)
    \vk_s
    \bigr]
    \boldsymbol{\rho}_s^\top .
\label{eq_app_output_token}
\end{align}
Define $\rmQ_{\gamma}$ by row $(\rmQ_{\gamma})_{r,:}=(\boldsymbol{\gamma}_r\odot\vq_r)^\top$ and define the causal score matrix
\begin{align}
    (\rmA_{qk})_{rs}
    =
    \mathbf{1}_{r\ge s}
    \vq_r^\top
    \operatorname{Diag}(\boldsymbol{\gamma}_r/\boldsymbol{\gamma}_s)
    \vk_s .
\label{eq_app_aqk}
\end{align}
Let $\rmO$ contain rows $\vo_r^\top$. Stacking Eq.~\ref{eq_app_output_token} over the chunk gives
\begin{align}
    \rmO
    =
    \rmQ_{\gamma}\rmS_0
    +
    \rmA_{qk}
    \bigl(\rmU-\rmY\rmS_0\bigr),
\label{eq_app_chunk_output}
\end{align}
which is Eq.~\ref{eq_gdn2_chunk_output} in the main text.

\subsection{Row recurrences}
\label{app_row_recurrences}

The matrices $\rmY$ and $\rmU$ can also be written row by row. Let row $r$ of $\rmY$ be $\vy_r^\top$ and row $r$ of $\rmU$ be $\vu_r^\top$. Since $(\rmI+\rmT)\rmY=\bar{\rmE}$ and $(\rmI+\rmT)\rmU=\rmZ$,
\begin{align}
    \vy_r^\top
    &=
    \bar{\ve}_r^\top
    -
    \sum_{s<r}
    \bar{\ve}_r^\top\bar{\vk}_s \, \vy_s^\top,
\label{eq_app_y_row}\\
    \vu_r^\top
    &=
    \vz_r^\top
    -
    \sum_{s<r}
    \bar{\ve}_r^\top\bar{\vk}_s \, \vu_s^\top .
\label{eq_app_u_row}
\end{align}
Both auxiliaries solve the same lower triangular system with different right-hand sides. This is why the same WY inverse can be shared by the erase-side and write-side computations.

\subsection{Tied-gate reductions}
\label{app_reductions}

If $\vb_r=\beta_r\mathbf{1}_{d_k}$ and $\vw_r=\beta_r\mathbf{1}_{d_v}$, then $\ve_r=\beta_r\vk_r$ and $\vz_r=\beta_r\vv_r$. Equation~\ref{eq_app_gdn2_step} becomes the KDA update. If the decay is also tied as $\valpha_r=\alpha_r\mathbf{1}_{d_k}$, the recurrence becomes Gated DeltaNet. Thus KDA and Gated DeltaNet are recovered by tying the channel gates rather than by changing the algorithm.

The same reduction holds for the chunkwise form. Under the KDA tying, the definitions in Eq.~\ref{eq_app_gate_matrices} give
\begin{align}
    \bar{\ve}_r
    =
    \boldsymbol{\gamma}_r\odot(\beta_r\vk_r)
    =
    \beta_r(\boldsymbol{\gamma}_r\odot\vk_r)
    =
    \beta_r(\boldsymbol{\gamma}_r\odot\boldsymbol{\gamma}_r)\odot\bar{\vk}_r,
    \qquad
    \vz_r
    =
    \beta_r\vv_r .
\label{eq_app_tied_gate_factors}
\end{align}
Thus $\rmZ$ becomes a scalar row scaling of $\rmV$, while $\bar{\rmE}$ becomes the KDA decay-normalized erase factor. In general, $\bar{\rmE}$ is not a scalar row scaling of $\bar{\rmK}$ when the decay is channel-wise, since each key channel carries its own factor from $\boldsymbol{\gamma}_r$. Equations~\ref{eq_app_wy_definitions}, \ref{eq_app_chunk_state}, and \ref{eq_app_chunk_output} therefore reduce to the KDA chunk equations after substituting the tied factors in Eq.~\ref{eq_app_tied_gate_factors}. If the decay is further tied as in Gated DeltaNet, then $\boldsymbol{\gamma}_r=\gamma_r\mathbf{1}_{d_k}$ and the erase factor simplifies to $\bar{\ve}_r=\beta_r\gamma_r^2\bar{\vk}_r$, recovering the scalar-decay chunkwise form.

At the gradient level, if a thin wrapper sets $\vb_r=\beta_r\mathbf{1}_{d_k}$ and $\vw_r=\beta_r\mathbf{1}_{d_v}$, the scalar gradient is
\begin{align}
    \frac{\partial \mathcal{L}}{\partial \beta_r}
    =
    \left\langle
    \frac{\partial \mathcal{L}}{\partial \vb_r},
    \mathbf{1}_{d_k}
    \right\rangle
    +
    \left\langle
    \frac{\partial \mathcal{L}}{\partial \vw_r},
    \mathbf{1}_{d_v}
    \right\rangle .
\label{eq_app_beta_gradient}
\end{align}

\section{Backward derivation}
\label{app_backward}

We derive the vector-Jacobian products for one chunk using the notation of Appendix~\ref{app_wy}. Let upstream gradients be $\mathrm{d}\rmO$ for the chunk output and $\mathrm{d}\rmS_C$ for the end-of-chunk state. The forward equations are
\begin{align}
    \rmR &= \rmU-\rmY\rmS_0,
\label{eq_app_bwd_R}\\
    \rmO &= \rmQ_{\gamma}\rmS_0+\rmA_{qk}\rmR,
\label{eq_app_bwd_O}\\
    \rmS_C &= \operatorname{Diag}(\boldsymbol{\gamma}_C)\rmS_0+\rmK_{\mathrm{tail}}^\top\rmR,
\label{eq_app_bwd_S}\\
    \rmU &= \rmA\rmZ,
    \qquad
    \rmY = \rmA\bar{\rmE},
    \qquad
    \rmA=(\rmI+\rmT)^{-1},
    \qquad
    \rmT=\operatorname{tril}(\bar{\rmE}\bar{\rmK}^\top,-1).
\label{eq_app_bwd_core}
\end{align}

\subsection{Output and state paths}
\label{app_bwd_paths}

From Eq.~\ref{eq_app_bwd_O},
\begin{align}
    \mathrm{d}\rmA_{qk}
    &\mathrel{+}=
    \mathrm{d}\rmO\,\rmR^\top,
\label{eq_app_daqk_from_output}\\
    \mathrm{d}\rmR
    &\mathrel{+}=
    \rmA_{qk}^\top\mathrm{d}\rmO,
\label{eq_app_dR_from_output}\\
    \mathrm{d}\rmQ_{\gamma}
    &\mathrel{+}=
    \mathrm{d}\rmO\,\rmS_0^\top,
\label{eq_app_dQg_from_output}\\
    \mathrm{d}\rmS_0
    &\mathrel{+}=
    \rmQ_{\gamma}^\top\mathrm{d}\rmO .
\label{eq_app_dS_from_output}
\end{align}
The causal mask is applied to $\mathrm{d}\rmA_{qk}$.

From Eq.~\ref{eq_app_bwd_S},
\begin{align}
    \mathrm{d}\rmR
    &\mathrel{+}=
    \rmK_{\mathrm{tail}}\mathrm{d}\rmS_C,
\label{eq_app_dR_from_state}\\
    \mathrm{d}\rmK_{\mathrm{tail}}
    &\mathrel{+}=
    \rmR\,\mathrm{d}\rmS_C^\top,
\label{eq_app_dKtail_from_state}\\
    \mathrm{d}\rmS_0
    &\mathrel{+}=
    \operatorname{Diag}(\boldsymbol{\gamma}_C)\mathrm{d}\rmS_C,
\label{eq_app_dS_from_state}\\
    \mathrm{d}\boldsymbol{\gamma}_C
    &\mathrel{+}=
    \mathrm{rowsum}\bigl(\mathrm{d}\rmS_C\odot\rmS_0\bigr).
\label{eq_app_dgammaC_from_state}
\end{align}
Here $\mathrm{rowsum}$ sums over the value dimension and returns a $d_k$-dimensional vector.

The residual relation in Eq.~\ref{eq_app_bwd_R} gives
\begin{align}
    \mathrm{d}\rmU
    &\mathrel{+}=
    \mathrm{d}\rmR,
\label{eq_app_dU_from_R}\\
    \mathrm{d}\rmY
    &\mathrel{+}=
    -\mathrm{d}\rmR\,\rmS_0^\top,
\label{eq_app_dY_from_R}\\
    \mathrm{d}\rmS_0
    &\mathrel{+}=
    -\rmY^\top\mathrm{d}\rmR .
\label{eq_app_dS_from_R}
\end{align}

\subsection{Gate-aware WY inverse path}
\label{app_bwd_wy}

The two auxiliary products yield
\begin{align}
    \mathrm{d}\rmA
    &\mathrel{+}=
    \mathrm{d}\rmU\,\rmZ^\top,
    &
    \mathrm{d}\rmZ
    &\mathrel{+}=
    \rmA^\top\mathrm{d}\rmU,
\label{eq_app_bwd_U}\\
    \mathrm{d}\rmA
    &\mathrel{+}=
    \mathrm{d}\rmY\,\bar{\rmE}^\top,
    &
    \mathrm{d}\bar{\rmE}
    &\mathrel{+}=
    \rmA^\top\mathrm{d}\rmY .
\label{eq_app_bwd_Y}
\end{align}
Equations~\ref{eq_app_bwd_U} and \ref{eq_app_bwd_Y} are the gate-aware accumulation emphasized in Section~\ref{subsec_gate_aware_backward}. Since $\rmZ=\rmW\odot\rmV$ and $\bar{\rmE}=\boldsymbol{\gamma}\odot(\rmB\odot\rmK)$, the gates must appear inside the products that accumulate $\mathrm{d}\rmA$. A scalar post-scale is correct only in the tied-gate case.

For the inverse,
\begin{align}
    \mathrm{d}\rmT
    =
    -\operatorname{tril}
    \bigl(
    \rmA^\top\mathrm{d}\rmA\rmA^\top,
    -1
    \bigr).
\label{eq_app_dT}
\end{align}
The construction of $\rmT$ gives
\begin{align}
    \mathrm{d}\bar{\rmE}
    &\mathrel{+}=
    \mathrm{d}\rmT\,\bar{\rmK},
\label{eq_app_dE_from_T}\\
    \mathrm{d}\bar{\rmK}
    &\mathrel{+}=
    \mathrm{d}\rmT^\top\bar{\rmE}.
\label{eq_app_dKbar_from_T}
\end{align}
Only the strictly lower triangular part of $\mathrm{d}\rmT$ is used.

\subsection{Scores and tail keys}
\label{app_bwd_scores}

The score matrix satisfies $\rmA_{qk}=\operatorname{tril}(\rmQ_{\gamma}\bar{\rmK}^\top)$. Therefore
\begin{align}
    \mathrm{d}\rmQ_{\gamma}
    &\mathrel{+}=
    \mathrm{d}\rmA_{qk}\,\bar{\rmK},
\label{eq_app_dQg_from_scores}\\
    \mathrm{d}\bar{\rmK}
    &\mathrel{+}=
    \mathrm{d}\rmA_{qk}^\top\rmQ_{\gamma}.
\label{eq_app_dKbar_from_scores}
\end{align}
The tail key matrix satisfies $(\rmK_{\mathrm{tail}})_{r}=\boldsymbol{\gamma}_C\odot\bar{\vk}_r$. Hence
\begin{align}
    \mathrm{d}\bar{\rmK}
    &\mathrel{+}=
    \mathrm{d}\rmK_{\mathrm{tail}}\odot\boldsymbol{\gamma}_C,
\label{eq_app_dKbar_from_tail}\\
    \mathrm{d}\boldsymbol{\gamma}_C
    &\mathrel{+}=
    \sum_{r=1}^{C}
    \mathrm{d}(\rmK_{\mathrm{tail}})_{r}
    \odot
    \bar{\vk}_r .
\label{eq_app_dgammaC_from_tail}
\end{align}

\subsection{Elementwise gates and cumulative decay}
\label{app_bwd_elementwise}

The write-side relation $\rmZ=\rmW\odot\rmV$ gives
\begin{align}
    \mathrm{d}\rmW
    &\mathrel{+}=
    \mathrm{d}\rmZ\odot\rmV,
    &
    \mathrm{d}\rmV
    &\mathrel{+}=
    \mathrm{d}\rmZ\odot\rmW .
\label{eq_app_dW_dV}
\end{align}
The erase-side relation $\bar{\rmE}=\boldsymbol{\gamma}\odot(\rmB\odot\rmK)$ gives
\begin{align}
    \mathrm{d}\rmB
    &\mathrel{+}=
    \mathrm{d}\bar{\rmE}\odot\boldsymbol{\gamma}\odot\rmK,
\label{eq_app_dB_from_E}\\
    \mathrm{d}\rmK
    &\mathrel{+}=
    \mathrm{d}\bar{\rmE}\odot\boldsymbol{\gamma}\odot\rmB,
\label{eq_app_dK_from_E}\\
    \mathrm{d}\boldsymbol{\gamma}
    &\mathrel{+}=
    \mathrm{d}\bar{\rmE}\odot\rmB\odot\rmK .
\label{eq_app_dgamma_from_E}
\end{align}
The normalized keys and queries are
\begin{align}
    \bar{\rmK} = \boldsymbol{\gamma}^{-1}\odot\rmK,
    \qquad
    \rmQ_{\gamma} = \boldsymbol{\gamma}\odot\rmQ .
\label{eq_app_Kbar_Qg}
\end{align}
Their vector-Jacobian products are
\begin{align}
    \mathrm{d}\rmK
    &\mathrel{+}=
    \mathrm{d}\bar{\rmK}\odot\boldsymbol{\gamma}^{-1},
\label{eq_app_dK_from_Kbar}\\
    \mathrm{d}\boldsymbol{\gamma}
    &\mathrel{+}=
    -\mathrm{d}\bar{\rmK}\odot\rmK\odot\boldsymbol{\gamma}^{-2},
\label{eq_app_dgamma_from_Kbar}\\
    \mathrm{d}\rmQ
    &\mathrel{+}=
    \mathrm{d}\rmQ_{\gamma}\odot\boldsymbol{\gamma},
\label{eq_app_dQ_from_Qg}\\
    \mathrm{d}\boldsymbol{\gamma}
    &\mathrel{+}=
    \mathrm{d}\rmQ_{\gamma}\odot\rmQ .
\label{eq_app_dgamma_from_Qg}
\end{align}
Finally, $\boldsymbol{\gamma}_r=\exp(\boldsymbol{G}_r)$ and $\boldsymbol{G}_r=\sum_{i\le r}\boldsymbol{g}_i$. Therefore
\begin{align}
    \mathrm{d}\boldsymbol{G}_r
    =
    \mathrm{d}\boldsymbol{\gamma}_r
    \odot
    \boldsymbol{\gamma}_r,
    \qquad
    \mathrm{d}\boldsymbol{g}_i
    =
    \sum_{r\ge i}
    \mathrm{d}\boldsymbol{G}_r .
\label{eq_app_dg_reverse_cumsum}
\end{align}
In implementation this is a reverse cumulative sum over the chunk.

\subsection{Why scalar post-scaling is invalid}
\label{app_no_post_scaling}

In KDA, a scalar $\beta_r$ multiplies both the value right hand side and the erase right hand side. For one row, the contribution to $\mathrm{d}\rmA$ from the write auxiliary can be factored as
\begin{align}
    \mathrm{d}\vu_r\,(\beta_s\vv_s)^\top
    =
    \beta_s\,\mathrm{d}\vu_r\vv_s^\top .
\label{eq_app_scalar_factor}
\end{align}
The scalar factor can be applied after the dot product. Gated DeltaNet-2 replaces $\beta_s\vv_s$ by $\vw_s\odot\vv_s$. Since $\vw_s$ is a different diagonal operator for every row, there is no row scalar or column scalar that can recover
\begin{align}
    \mathrm{d}\vu_r\,(\vw_s\odot\vv_s)^\top
\label{eq_app_vector_factor_write}
\end{align}
from $\mathrm{d}\vu_r\vv_s^\top$. The erase side has the same issue with $\vb_s\odot\vk_s$. The gates must be baked into the dot products in Eq.~\ref{eq_app_bwd_U} and Eq.~\ref{eq_app_bwd_Y}.

\section{Layer and kernel implementation}
\label{app_implementation}

This appendix records the implementation choices needed to reproduce Gated DeltaNet-2. The main text keeps the Triton details brief. Here we describe the computation at the level of kernels and tensor shapes.

\subsection{Layer parameterization}
\label{app_layer}

The layer computes short-convolutional projections for $\vq$, $\vk$, and $\vv$, followed by head reshaping. The erase and write gates are produced by independent projections,
\begin{align}
    \vb = \sigma(\operatorname{Proj}_{b}(\vx)),
    \qquad
    \vw = \sigma(\operatorname{Proj}_{w}(\vx)).
\label{eq_app_layer_gates}
\end{align}
The erase projection has shape $d_{\mathrm{model}}\to H d_k$. The write projection has shape $d_{\mathrm{model}}\to H_v d_v$. If grouped value attention is used with $H_v>H$, the key-side tensors $\vq$, $\vk$, the log-decay tensor $\boldsymbol{g}$, and $\vb$ are repeated across the value-head group, while $\vv$ and $\vw$ already live on the value-head axis.

The log-decay is computed outside the kernel in fp32,
\begin{align}
    \boldsymbol{g}_t
    =
    -\exp(\mathbf{a})\odot
    \operatorname{softplus}(\operatorname{Proj}_{f}(\vx_t)+\boldsymbol{\delta}).
\label{eq_app_layer_decay}
\end{align}
The vector $\mathbf{a}$ is stored per key head and broadcast across the $d_k$ channels of that head. The bias $\boldsymbol{\delta}$ is stored per key channel. The kernel consumes $\boldsymbol{g}_t$ directly and forms the local cumulative sums in Eq.~\ref{eq_app_gamma}.

If negative eigenvalues are enabled, only the erase gate is scaled by $2$. This changes $\vb_t\in[0,1]^{d_k}$ into $\vb_t\in[0,2]^{d_k}$. The write gate remains in $[0,1]^{d_v}$.

\subsection{Forward kernels}
\label{app_forward_kernels}

The chunk size is fixed to $C=64$. Each chunk is processed by the following steps.

\paragraph{Intra-chunk products}
The first kernel forms the causal score matrix $\rmA_{qk}$ and the strictly lower matrix $\rmT$. The Gated DeltaNet-2 specific computation is the row factor of $\rmT$,
\begin{align}
    T_{rs}
    =
    \bar{\ve}_r^\top\bar{\vk}_s
    =
    (\boldsymbol{\gamma}_r\odot\vb_r\odot\vk_r)^\top
    (\boldsymbol{\gamma}_s^{-1}\odot\vk_s)
    \qquad
    s<r .
\label{eq_app_kernel_T}
\end{align}
Thus the erase gate is multiplied into the key tile before the dot product. The score matrix $\rmA_{qk}$ is unchanged apart from the same decay-normalized key factors.

\paragraph{WY solve}
The second kernel solves $\rmA=(\rmI+\rmT)^{-1}$ by forward substitution. It then exposes the same lower triangular inverse to both right hand sides. This is the compact WY step used in Eq.~\ref{eq_app_wy_definitions}.

\paragraph{Auxiliary construction}
The third kernel builds
\begin{align}
    \rmU = \rmA(\rmW\odot\rmV),
    \qquad
    \rmY = \rmA\bar{\rmE}.
\label{eq_app_kernel_aux}
\end{align}
The implementation stores $\rmY$ using the historical buffer name \texttt{w} because KDA used the same buffer for its erase-side auxiliary; this buffer is not the write-gate matrix $\rmW$. The mathematical role is $\rmY$ throughout this paper.

\paragraph{State and output}
The inter-chunk state recurrence consumes $\rmK_{\mathrm{tail}}$, $\rmY$, and $\rmU$, and applies Eq.~\ref{eq_app_chunk_state}. The output kernel consumes $\rmQ_{\gamma}$, $\rmA_{qk}$, and $\rmR=\rmU-\rmY\rmS_0$, and applies Eq.~\ref{eq_app_chunk_output}. These two kernels do not depend on how the gate factors were produced, so they share the same matrix shapes as KDA.

\subsection{Backward kernels}
\label{app_backward_kernels}

The backward pass mirrors Appendix~\ref{app_backward}.

\paragraph{Output vector-Jacobian product}
The first backward kernel computes $\mathrm{d}\rmA_{qk}$ and the output-path contribution to $\mathrm{d}\rmR$ from Eq.~\ref{eq_app_daqk_from_output} and Eq.~\ref{eq_app_dR_from_output}. It is structure-equivalent to the KDA output vector-Jacobian product because it only sees $\rmR$.

\paragraph{State vector-Jacobian product}
The second backward kernel propagates $\mathrm{d}\rmS_C$ through Eq.~\ref{eq_app_bwd_S}. It is also structure-equivalent to the KDA state vector-Jacobian product because it consumes $\rmK_{\mathrm{tail}}$, $\rmY$, and $\rmU$ as already formed tensors.

\paragraph{Gate-aware WY vector-Jacobian product}
The third backward kernel implements Eq.~\ref{eq_app_bwd_U}, Eq.~\ref{eq_app_bwd_Y}, and Eq.~\ref{eq_app_dT}. This is the main Gated DeltaNet-2 specific kernel. It accumulates $\mathrm{d}\rmA$ with $\rmZ^\top=(\rmW\odot\rmV)^\top$ and $\bar{\rmE}^\top=(\boldsymbol{\gamma}\odot\rmB\odot\rmK)^\top$. It also emits the direct gradients
\begin{align}
    \mathrm{d}\rmW
    =
    \mathrm{d}\rmZ\odot\rmV,
    \qquad
    \mathrm{d}\rmV
    =
    \mathrm{d}\rmZ\odot\rmW,
    \qquad
    \mathrm{d}\rmB
    \mathrel{+}=
    \mathrm{d}\bar{\rmE}\odot\boldsymbol{\gamma}\odot\rmK .
\label{eq_app_kernel_gate_grads}
\end{align}
The erase-gate gradient has shape $B\times T\times H\times d_k$. The write-gate gradient has shape $B\times T\times H_v\times d_v$.

\paragraph{Intra-chunk vector-Jacobian product}
The fourth backward kernel propagates through $\rmA_{qk}$ and $\rmT$. It adds the remaining contributions to $\mathrm{d}\rmQ$, $\mathrm{d}\rmK$, $\mathrm{d}\rmB$, and $\mathrm{d}\boldsymbol{g}$. The dependence on the cumulative decay is reduced by a reverse cumulative sum, as in Eq.~\ref{eq_app_dg_reverse_cumsum}.

\subsection{Autotuning and hardware dispatch}
\label{app_autotune}

The fused WY backward kernel uses the same matrix shapes as the forward solve, but it has a denser set of live accumulators because it emits gradients for $\rmB$ and $\rmW$. On Hopper GPUs we restrict the warp search for this kernel to two and four warps, since the eight-warp schedule can trigger a Triton WGMMA layout assertion for the $64\times64$ accumulator. On Ampere GPUs the full search space is retained. This restriction changes only the schedule, not the mathematical operation.

\subsection{Recurrent decoding kernel}
\label{app_recurrent_kernel}

A forward-only recurrent kernel is provided for autoregressive decoding at short sequence lengths. It applies Eq.~\ref{eq_app_gdn2_step} token by token. The kernel keeps the state in fp32, multiplies it by $\exp(\boldsymbol{g}_t)$, reads the decayed state along $\vb_t\odot\vk_t$, writes $\vw_t\odot\vv_t$ along $\vk_t$, and returns $\rmS_t^\top\vq_t$. Training uses the chunk kernel.

\subsection{Variable-length sequences}
\label{app_varlen}

Packed variable-length batches are represented with cumulative sequence lengths. The chunk index construction resets the recurrent state at every sequence boundary. The same layout is used by the chunk forward, the chunk backward, and the recurrent decoding kernel. Padding is removed before the layer and restored after the output projection.

\section{Numerical details and verification}
\label{app_numerics}

\subsection{Decay precision}
\label{app_decay_precision}

The decay gate in Eq.~\ref{eq_app_layer_decay} is computed in explicit fp32 before entering the kernels. This is important because the local cumulative sum $\boldsymbol{G}_r=\sum_{i\le r}\boldsymbol{g}_i$ is a path-length-dependent quantity. A low precision mantissa can perturb long products of decays even when each tokenwise gate is small. The kernels therefore receive the log-decay tensor and only compute local cumulative sums and exponentials.

\subsection{Query and key normalization}
\label{app_qk_norm}

Queries and keys are L2-normalized per head before the recurrent update. With normalized keys, $\vk_t\vk_t^\top$ is a projector in the tied-gate limit. In Gated DeltaNet-2, the erase factor is asymmetric, but normalization still stabilizes the scale of both $\rmA_{qk}$ and $\rmT$. The backward applies the standard L2-normalization vector-Jacobian product.

\subsection{State and accumulator dtypes}
\label{app_dtype}

The recurrent state is stored in fp32 across chunks and during recurrent decoding. Matrix multiplication accumulators use fp32. The layer output is cast back to the model dtype at the kernel boundary. The WY auxiliaries may be stored in the model dtype after fp32 accumulation, since they are recomputed when the memory-saving training path is used.

\subsection{WY solve precision}
\label{app_solve_precision}

The triangular solve for $\rmA=(\rmI+\rmT)^{-1}$ is the most precision-sensitive part of the chunk computation. Errors in this solve are propagated through dependent forward-substitution steps. The implementation therefore exposes an explicit precision flag for the solve. The conservative IEEE fp32 path is used when required by the hardware check, while the remaining matrix products can use the faster tensor-core path.

\subsection{Initialization and output gate}
\label{app_init_output_gate}

All linear layers are initialized with Xavier uniform weights and gain $2^{-2.5}$. Biases are initialized to zero when present. After the attention computation, the output is passed through an RMSNorm and SiLU gate before the final output projection. These choices match the training recipe used for the Gated DeltaNet family and keep the early recurrent state magnitudes controlled.

\subsection{Correctness checks}
\label{app_correctness}

We verified the chunkwise forward against a tokenwise recurrent reference for random configurations covering different sequence lengths, head counts, key dimensions, value dimensions, initial states, packed layouts, and dtypes. We verified the backward against autograd through the recurrent reference. In fp64 reference tests, gradients for $\rmQ$, $\rmK$, $\rmV$, $\rmB$, $\rmW$, the log-decay, and the initial state agree to machine precision. In production fp32, differences are at the expected tensor-core accumulation noise level. In bfloat16, the error follows the bfloat16 mantissa.

\section{Experimental settings}
\label{app_experimental_settings}

\subsection{Training}

We evaluate Gated DeltaNet-2 against a Transformer baseline and recent recurrent architectures, including Mamba-2 \citep{pmlr-v235-dao24a}, Gated DeltaNet \citep{yang2025gated}, Kimi Delta Attention (KDA) \citep{team2025kimi}, and Mamba-3 \citep{lahoti2026mamba3}. For each recurrent architecture, we train a recurrent-only model and a hybrid model. The hybrid model follows Section~\ref{subsec_block_design}, using the same recurrent token mixer together with sliding-window attention (SWA) under the same residual block structure. For Mamba-3, we evaluate both SISO and MIMO variants. The Mamba-3 MIMO model uses rank $R=4$.

For fair recurrent comparisons, we match both parameter count and main recurrent state size. Gated DeltaNet, KDA, and Gated DeltaNet-2 use $H=16$ heads with $d_k=128$ and $d_v=128$, giving a per-layer recurrent state of
\begin{align}
    H d_k d_v
    =
    16 \cdot 128 \cdot 128
    =
    262{,}144
\end{align}
floats per batch element. Since $d_{\mathrm{model}}=2048$, this equals $128d_{\mathrm{model}}$. For Mamba-2 and Mamba-3, we use expansion factor $2$ and head dimension $64$, and set $d_{\mathrm{state}}=64$. Their main recurrent state size is therefore
\begin{align}
    (2d_{\mathrm{model}})d_{\mathrm{state}}
    =
    4096 \cdot 64
    =
    262{,}144 .
\end{align}
Mamba-3 MIMO keeps the same main recurrent state size as the SISO variant while adding the rank-$R$ MIMO parameterization.

Unless stated otherwise, all models have 1.3B parameters and are trained on 100B tokens sampled from FineWeb-Edu \citep{penedo2024fineweb}. We use AdamW with peak learning rate $4\times10^{-4}$, weight decay $0.1$, and gradient clipping at $1.0$. The learning rate follows cosine annealing with a 1B-token warm-up. The global batch size is 0.5M tokens. The training sequence length is 4K tokens. Hybrid models use a 2K SWA window.

\subsection{Evaluation}

\paragraph{Language modeling and common-sense reasoning}
We use the evaluation suite commonly adopted for pretrained recurrent language models \citep{gu_mamba_2023}. Language modeling quality is measured by perplexity on WikiText \citep[Wiki.][]{merity2016pointer} and LAMBADA \citep[LMB.][]{paperno_lambada_2016}. For zero-shot transfer, we report LAMBADA accuracy together with PIQA \citep{bisk2020piqa}, HellaSwag \citep[Hella.][]{zellers2019hellaswag}, WinoGrande \citep[Wino.][]{sakaguchi2021winogrande}, ARC-Easy and ARC-Challenge \citep[ARC-e and ARC-c][]{arc-ce}, OpenBookQA \citep[OBQA][]{openbookqa}, Social IQa \citep[SIQA][]{sap2019social}, and BoolQ \citep{clark2019boolq}. This mix covers next-token prediction, physical and social reasoning, commonsense completion, and elementary science QA.

\paragraph{In-context retrieval}
We evaluate retrieval in both controlled synthetic settings and real-data settings. For synthetic retrieval, we use Single Needle-In-A-Haystack (S-NIAH) and Multi-Key Needle-In-A-Haystack (MK-NIAH) tasks from RULER \citep{hsieh2024ruler}. The S-NIAH suite contains three progressively harder cases. S-NIAH-1 is passkey retrieval, S-NIAH-2 asks for a numerical needle, and S-NIAH-3 asks for a word-based needle. We additionally evaluate MK-NIAH-1 where several distractor key-value pairs are present and the model must return the value associated with one requested key.

For real-world retrieval, we follow \citet{arora-2024-jrt}. The suite includes SWDE \citep{lockard_openceres_2019} for structured relation extraction from HTML, FDA \citep{arora_language_2023} for key-value retrieval from PDFs, and question-answering datasets including SQuAD \citep{rajpurkar_know_2018}, TriviaQA \citep{JoshiTriviaQA2017}, DROP \citep{dua2019drop}, and Natural Questions \citep{47761}.

\subsection{Large-scale global-attention hybrid}
\label{app_large_scale_setting}

The scaling experiment in Table~\ref{tab_scaling_hybrid} uses a Nemotron-3 Nano-style hybrid latent-MoE architecture~\citep{nvidia2025nemotron3,nvidia2025nemotron3nano}, in which recurrent mixers and MoE layers are interleaved with a small number of full global-attention layers. Each model has 31.6B total parameters and 3.2B active parameters and is trained for 300B tokens. The global-attention layers and their placement, the recurrent-state size, the active parameter count, and the training recipe are identical across variants, and only the recurrent mixer is changed among Mamba-2, KDA, and Gated DeltaNet-2. Since the mixers differ slightly in parameter count, the FFN intermediate size is adjusted by a small amount so that all variants are iso-parameter.

The evaluation covers four categories. Knowledge is measured with MMLU-Pro \citep{wang2024mmlupro} and MMLU-Str.\ \citep{hendrycks2021mmlu}. 
Code is measured with HumanEval \citep{chen2021codex} and MBPP \citep{austin2021mbpp}. Math is measured with GSM8K \citep{cobbe2021gsm8k} and MATH-500 \citep{hendrycks2021math,lightman2024verify}. Commonsense understanding is measured with RACE \citep{lai2017race}, ARC-Challenge \citep{arc-ce}, HellaSwag \citep{zellers2019hellaswag}, and WinoGrande \citep{sakaguchi2021winogrande}. Each category score is the mean of its benchmarks, and the overall score is the equal average of the four category scores.


\section{Extended related work}
\label{app_related_extended}

Section~\ref{sec_related_work} positions Gated DeltaNet-2 among efficient sequence models. This appendix expands that discussion along the axis most relevant to our contribution, the use of several gates to control a recurrent memory, and relates the model to the associative-memory view of recent work.

\paragraph{Gated recurrent memories}
The principle of separating memory-control decisions into several gates dates back to the LSTM, whose elementwise input and forget gates act on a vector-valued cell state~\citep{hochreiter_long_1997}. The mLSTM cell of xLSTM introduces a matrix memory, but still updates it by a gated covariance accumulation~\citep{beck2024xlstm}. Modern linear attention and linear RNNs make the decay data-dependent and often vector-valued. Gated linear attention applies data-dependent vector gating to a matrix-state linear attention layer while retaining an additive key-value write~\citep{yang_gated_2023}, RWKV uses channel-wise decay and its RWKV-5 and RWKV-6 successors adopt matrix-valued states with dynamic vector decay~\citep{peng_rwkv_2023,peng_eagle_2024}, HGRN2 expands the state with outer products under vector forget gates~\citep{qin2024hgrn2}, Mamba-2 uses a scalar data-dependent decay~\citep{pmlr-v235-dao24a}, and KDA makes the decay of a delta-rule layer channel-wise~\citep{team2025kimi}. None of the gated accumulators in this family performs the read-before-write delta correction of Gated DeltaNet-2,
\begin{align}
    \rmS_t
    =
    \bar{\rmS}_t
    +
    \vk_t
    \bigl[
    (\vw_t\odot\vv_t)
    -
    \bar{\rmS}_t^\top(\vb_t\odot\vk_t)
    \bigr]^\top,
    \qquad
    \bar{\rmS}_t=\rmD_t\rmS_{t-1}.
\label{eq_app_related_delta}
\end{align}
Our contribution is therefore not the generic use of multiple or vector-valued gates. It is that the two gates act on different axes of a queryable matrix memory inside a delta-rule edit. The erase gate $\vb_t$ controls the key-side direction along which an existing association is read and erased, while the write gate $\vw_t$ independently controls which value channels are committed to memory.

\paragraph{Delta-rule and test-time-learning memories}
DeltaNet, Gated DeltaNet, and KDA, discussed in Section~\ref{sec_preliminary}, all use a single scalar step size $\beta_t$ for the delta edit. Two models relax this scalar. RWKV-7 generalizes the delta rule with vector-valued in-context learning rates and a removal key that is decoupled from the replacement key, which makes it structurally the closest prior model to ours at the level of state evolution~\citep{peng2025rwkv7}. FG$^2$-GDN, concurrent with this work, makes $\beta_t$ channel-wise, and its FG$^2$-GDN+ variant uses separate channel-wise scalings for keys and values~\citep{sun2026fg2gdn}. Appendix~\ref{app_rwkv7_fg2gdn} gives the formulation-level differences between these models and Gated Delta Rule-2 and reports a matched comparison. Beyond the step size, DeltaProduct composes several Householder transformations per token to improve state tracking~\citep{siems2025deltaproduct}, and \citet{Grazzi2024UnlockingSI} show that allowing negative eigenvalues in the state transition unlocks state tracking in linear RNNs. Gated DeltaNet-2 supports the corresponding expanded erase range, and Appendix~\ref{app_state_tracking} evaluates it on the DeltaProduct permutation tasks. The fast-weight interpretation of Section~\ref{sec_method} follows the view of recurrent layers as online learners~\citep{linear-xmr-fastweight,Irie2022TheDF,ttt,longhorn}.

\paragraph{The Miras view}
The Miras framework organizes sequence models through four design choices, the associative-memory architecture, the attentional-bias objective, the retention gate, and the memory-learning algorithm~\citep{behrouz2025miras}. Under this taxonomy, Gated DeltaNet-2 uses a matrix associative memory, a regression-based delta objective, Eq.~\ref{eq_gdn2_online_objective}, a channel-wise diagonal retention gate $\rmD_t$, and a one-step online memory update. The retention-gate axis most directly describes how the previous memory is preserved or forgotten, which corresponds to $\rmD_t$ in Gated DeltaNet-2. Our model focuses on the finer question of how the subsequent active delta edit is factorized,
\begin{align}
    \ve_t=\vb_t\odot\vk_t,
    \qquad
    \vz_t=\vw_t\odot\vv_t,
    \qquad
    \rmS_t
    =
    \rmD_t\rmS_{t-1}
    +
    \vk_t
    \bigl(
    \vz_t-(\rmD_t\rmS_{t-1})^\top\ve_t
    \bigr)^\top,
\end{align}
so that $\vb_t$ independently controls the key-side read and erase direction while $\vw_t$ controls the value-side commit target. This is a finer modeling choice within the Miras framework rather than an alternative framework.

\paragraph{Hybrid architectures}
Recurrent mixers are commonly combined with attention, either with sliding windows as in Griffin and Samba~\citep{de_griffin_2024,ren2024samba} or with a small number of full global-attention layers as in Jamba and the Nemotron family~\citep{Lieber2024JambaAH,nvidia2025nemotron3,nvidia2025nemotron3nano}. Our 1.3B hybrids follow the first pattern, and the scaling experiment of Table~\ref{tab_scaling_hybrid} follows the second.

\section{Comparison with RWKV-7 and FG$^2$-GDN+}
\label{app_rwkv7_fg2gdn}

RWKV-7~\citep{peng2025rwkv7} and the concurrent FG$^2$-GDN+~\citep{sun2026fg2gdn} both relax the single scalar step size of the delta rule, so it is natural to ask what Gated Delta Rule-2 adds. This appendix compares the three updates at the level of their equations and then reports a matched empirical comparison.

\subsection{Formulation-level differences}
\label{app_formulation_differences}

\paragraph{FG$^2$-GDN+}
FG$^2$-GDN+ predicts a channel-wise key step size $\boldsymbol{\beta}^{k}_t\in\mathbb{R}^{d_k}$ and a channel-wise value step size $\boldsymbol{\beta}^{v}_t\in\mathbb{R}^{d_v}$ and folds their square roots into the key and value,
\begin{align}
    \hat{\vk}_t
    =
    \sqrt{\boldsymbol{\beta}^{k}_t}\odot\vk_t,
    \qquad
    \tilde{\vv}_t
    =
    \sqrt{\boldsymbol{\beta}^{v}_t}\odot\vv_t,
    \qquad
    \rmS_t
    =
    \bigl(\rmI-\hat{\vk}_t\hat{\vk}_t^\top\bigr)\rmD_t\rmS_{t-1}
    +
    \hat{\vk}_t\tilde{\vv}_t^\top .
\label{eq_app_fg2gdn_plus}
\end{align}
The same vector $\hat{\vk}_t$ controls key-side erasure and key-side writing. Weakening erasure on key channel $i$ therefore also weakens every write into row $i$ of the state. The value gate $\boldsymbol{\beta}^{v}_t$ cannot compensate, because it scales value channels, the columns of $\rmS_t$, rather than key channels, the rows. Gated Delta Rule-2 instead uses
\begin{align}
    \rmS_t
    =
    \bigl(\rmI-\vk_t(\vb_t\odot\vk_t)^\top\bigr)\rmD_t\rmS_{t-1}
    +
    \vk_t(\vw_t\odot\vv_t)^\top,
\end{align}
where $\vb_t$ affects only the read and erase direction, writing remains along the unmodified key $\vk_t$, and $\vw_t$ independently selects the committed value channels. Thus FG$^2$-GDN+ decouples the erase magnitude from value-side scaling, but retains a hard key-side coupling between erasing and writing that Gated Delta Rule-2 removes.

\paragraph{RWKV-7}
RWKV-7 removes old content with a token-dependent channel-wise strength $\va_t$ and writes the new value along a replacement key that is a soft function of the same signal,
\begin{align}
    \tilde{\vk}_t
    =
    \vk_t\odot\operatorname{lerp}(\vone,\va_t,\boldsymbol{\xi})
    =
    \vk_t\odot\bigl[\vone+(\va_t-\vone)\odot\boldsymbol{\xi}\bigr],
\label{eq_app_rwkv7_key}
\end{align}
where $\boldsymbol{\xi}$, denoted $\alpha$ in RWKV-7, is a learned token-independent channel-wise replacement-rate booster. The same $\va_t$ therefore influences both how much content is removed and the key used to write new content, with $\boldsymbol{\xi}$ controlling the strength of this coupling. Setting $\boldsymbol{\xi}$ near zero on a channel removes the coupling, but only statically. RWKV-7 cannot choose the coupling independently per token, and it has no independently predicted gate that directly scales value channels in the state write. Gated Delta Rule-2 predicts independent token-dependent $\vb_t$ and $\vw_t$ by construction and removes the coupling across the two axes of its matrix memory.

\subsection{Matched comparison}
\label{app_matched_comparison}

To compare the three updates empirically, we implemented the RWKV-7 and FG$^2$-GDN+ state updates in the same surrounding block as Gated DeltaNet-2 (Section~\ref{subsec_block_design}) and matched the parameter count, the recurrent-state size, the 1.3B scale, the 100B-token FineWeb-Edu budget, the 4K training length, and the training recipe of Section~\ref{sec_exp}. Table~\ref{tab_app_rwkv7_fg2gdn} reports three aggregates. Avg.\ ppl averages WikiText and LAMBADA perplexity, Common Avg.\ follows the average of Table~\ref{tab_commonsense_results}, and Recall Avg.\ averages the six real-world retrieval tasks of Table~\ref{tab_recall_results}. Results are the mean and sample standard deviation over three independent pretraining seeds. The single-run Gated DeltaNet-2 checkpoint of Tables~\ref{tab_commonsense_results} to~\ref{tab_recall_results} corresponds to the first of these seeds and closely matches the three-seed mean.

\begin{table}[H]
\centering
\small
\setlength{\tabcolsep}{8pt}
\begin{tabular}{l|ccc}
\toprule
\textbf{Model} & \textbf{Avg.\ ppl} $\downarrow$ & \textbf{Common Avg.} $\uparrow$ & \textbf{Recall Avg.} $\uparrow$ \\
\midrule
RWKV-7 & $14.27 \pm 0.05$ & $52.21 \pm 0.14$ & $28.54 \pm 0.17$ \\
FG$^2$-GDN+ & $14.18 \pm 0.04$ & $52.38 \pm 0.13$ & $28.73 \pm 0.16$ \\
Gated DeltaNet-2 & $\mathbf{13.66 \pm 0.03}$ & $\mathbf{53.10 \pm 0.10}$ & $\mathbf{29.87 \pm 0.12}$ \\
\bottomrule
\end{tabular}
\caption{
Matched comparison of RWKV-7, FG$^2$-GDN+, and Gated DeltaNet-2 in the recurrent-only 1.3B setting with 100B training tokens.
All models share the same block design, parameter count, recurrent-state size, and training recipe.
Values are mean $\pm$ sample standard deviation over three pretraining seeds.
}
\label{tab_app_rwkv7_fg2gdn}
\end{table}

Gated DeltaNet-2 outperforms both models on every aggregate, with margins well above the seed variation. Relative to FG$^2$-GDN+, it reduces average perplexity by 0.52 and improves the commonsense and recall averages by 0.72 and 1.14 points. Relative to RWKV-7, it reduces average perplexity by 0.61 and improves the two averages by 0.89 and 1.33 points. Since the three models differ only in how the delta edit is gated, these margins reflect the value of predicting the key-side erase direction and the value-side write target independently at every token.

\section{Additional ablations on the gate structure}
\label{app_additional_ablations}

Table~\ref{tab_ablation} tests whether the erase and write gates need channel-wise structure by scalarizing either gate. This appendix adds two controls that isolate other aspects of Gated Delta Rule-2: the independence of the two gates and the incremental contribution of write selectivity. Both use the recurrent-only 1.3B setting and the metrics of Table~\ref{tab_ablation}. As in Table~\ref{tab_ablation}, the tables report single-run values; the shared-gate variant of Section~\ref{app_tied_vs_decoupled} was additionally trained with the same seed and data order as the Gated DeltaNet-2 checkpoint of Table~\ref{tab_ablation}, so that comparison is paired.

\subsection{Tied versus decoupled channel-wise gates}
\label{app_tied_vs_decoupled}

Comparing Gated DeltaNet-2 with KDA changes two things at once. The delta gate becomes channel-wise, and the erase and write decisions become independent. The scalarized variants of Table~\ref{tab_ablation} isolate the first change. To isolate the second, we train a shared-vector-gate variant that retains both gate projections and ties only their token-dependent outputs,
\begin{align}
    \vg_t
    =
    \tfrac{1}{2}
    \bigl[
    \sigma(\mathbf{W}_b\vx_t)
    +
    \sigma(\mathbf{W}_w\vx_t)
    \bigr],
    \qquad
    \vb_t=\vw_t=\vg_t,
\label{eq_app_shared_gate}
\end{align}
with the recurrence
\begin{align}
    \rmS_t
    =
    \bigl(\rmI-\vk_t(\vg_t\odot\vk_t)^\top\bigr)\rmD_t\rmS_{t-1}
    +
    \vk_t(\vg_t\odot\vv_t)^\top .
\label{eq_app_shared_gate_recurrence}
\end{align}
Because $d_k=d_v$ in our configuration, the same channel-wise vector can be applied on both axes. The control preserves the parameter count, the gate projections, the recurrent-state size, the compute, and the training recipe of Gated DeltaNet-2, and removes only the independence between erasing and writing.

\begin{table}[H]
\centering
\scriptsize
\addtolength{\tabcolsep}{-1.5pt}
\begin{tabular}{l|cc|c|ccc|c}
\toprule
\textbf{Variant} & \textbf{Wiki.} & \textbf{LMB.} & \textbf{Common.} &
\textbf{S-NIAH-2} & \textbf{S-NIAH-3} & \textbf{MK-NIAH-1} & \textbf{Recall} \\
 & ppl $\downarrow$ & ppl $\downarrow$ & avg $\uparrow$ &
@4K $\uparrow$ & @2K $\uparrow$ & @4K $\uparrow$ & avg $\uparrow$ \\
\midrule
Shared vector gate, $\vb_t=\vw_t$ & 16.07 & 11.49 & 52.86 & 91.8 & 85.2 & 35.1 & 29.52 \\
Independent $\vb_t$, $\vw_t$ (Gated DeltaNet-2) & \textbf{15.90} & \textbf{11.41} & \textbf{53.11} & \textbf{93.0} & \textbf{89.8} & \textbf{37.8} & \textbf{29.88} \\
\bottomrule
\end{tabular}
\addtolength{\tabcolsep}{1.5pt}
\caption{
Tied versus decoupled channel-wise gates in the recurrent-only setting.
The shared-vector-gate variant keeps both gate projections and ties their outputs, so it matches Gated DeltaNet-2 in parameters, state size, and compute.
}
\label{tab_app_shared_gate}
\end{table}

Table~\ref{tab_app_shared_gate} shows that independent gates improve every metric, including 4.6 points on S-NIAH-3 and 2.7 points on MK-NIAH-1. Since vectorization, parameters, state size, compute, and training are fixed, the difference directly isolates the benefit of allowing key-side erasure and value-side writing to vary independently. The same control also rules out a capacity explanation of the gain over KDA. The shared-gate model retains both $\mathbf{W}_b$ and $\mathbf{W}_w$ and therefore has exactly the parameter count of Gated DeltaNet-2, yet it still trails on every reported metric.

\subsection{Erase versus write selectivity}
\label{app_erase_vs_write}

Table~\ref{tab_ablation} indicates that key-side erase selectivity accounts for most of the improvement at the current scale and context length. To isolate the additional contribution of write selectivity, Table~\ref{tab_app_write_selectivity} compares the variant that keeps $\vb_t$ channel-wise but scalarizes $\vw_t$ against the full model, using the single-run values of Table~\ref{tab_ablation}.

\begin{table}[H]
\centering
\scriptsize
\setlength{\tabcolsep}{6pt}
\begin{tabular}{l|c|ccc|c}
\toprule
\textbf{Variant} & \textbf{Common.} & \textbf{S-NIAH-2} & \textbf{S-NIAH-3} & \textbf{MK-NIAH-1} & \textbf{Recall} \\
 & avg $\uparrow$ & @4K $\uparrow$ & @2K $\uparrow$ & @4K $\uparrow$ & avg $\uparrow$ \\
\midrule
Channel-wise $\vb_t$, scalar $\vw_t$ & 52.79 & 92.1 & 84.6 & 35.2 & 29.51 \\
Channel-wise $\vb_t$ and $\vw_t$ (Gated DeltaNet-2) & \textbf{53.11} & \textbf{93.0} & \textbf{89.8} & \textbf{37.8} & \textbf{29.88} \\
\bottomrule
\end{tabular}
\caption{
Incremental contribution of write selectivity in the recurrent-only setting.
Both rows keep the erase gate channel-wise; the first scalarizes the write gate as in Table~\ref{tab_ablation}.
}
\label{tab_app_write_selectivity}
\end{table}

Erase selectivity is the larger contributor at the 4K training length, but write selectivity provides substantial additional gains on interference-heavy retrieval, improving S-NIAH-3 by 5.2 points and MK-NIAH-1 by 2.6 points. The smaller gains on the commonsense and recall averages indicate that this effect is concentrated in settings where several associations compete for limited recurrent memory. The two gates address complementary bottlenecks. The erase gate determines which components of an existing association should be revised, while the write gate acts as admission control over which value channels enter memory. We therefore expect erase selectivity to remain important for overwrite-heavy retrieval, while write selectivity should become more valuable as longer and more information-dense contexts increase the pressure to avoid storing irrelevant content.

\section{State tracking on permutation composition}
\label{app_state_tracking}

The main text focuses on retrieval, where fixed-state memory must manage interference. State tracking is a second capability that motivates recent recurrent models~\citep{Grazzi2024UnlockingSI,siems2025deltaproduct,lahoti2026mamba3}. We therefore evaluate Gated DeltaNet-2 on the permutation-composition benchmark of DeltaProduct~\citep{siems2025deltaproduct}, which builds on the group word problems of \citet{DBLP:conf/icml/MerrillPS24}. Each input is a sequence of elements of a finite group and the model must track their running composition. We use the symmetric groups $S_3$, $S_4$, and $S_5$ and the alternating group $A_5$. Following the DeltaProduct protocol, models are trained at length 128 and evaluated at length 512, so the task also measures length extrapolation. We match model depth, hidden size, recurrent-state size, and optimization across architectures and report the mean and standard deviation over three seeds.

DeltaProduct reports that the standard nonnegative transition range fails on these tasks. Gated DeltaNet, KDA, and Gated DeltaNet-2 therefore use their state-tracking-enabled gate range, $\beta_t\in[0,2]$ for the scalar-gated models and $\vb_t\in[0,2]^{d_k}$ for Gated DeltaNet-2 as described in Section~\ref{subsec_decoupling}, while Mamba-3 uses its data-dependent rotations~\citep{lahoti2026mamba3}. This expanded erase range is already supported by the Gated DeltaNet-2 kernels, and Table~\ref{tab_ablation} shows that it is neutral for language modeling and retrieval at 1.3B. We additionally report Gated DeltaNet-2 with the standard range $\vb_t\in[0,1]^{d_k}$ as a control.

\begin{table}[H]
\centering
\small
\setlength{\tabcolsep}{7pt}
\begin{tabular}{l|cccc}
\toprule
\textbf{Model} & $\boldsymbol{S_3}$ $\uparrow$ & $\boldsymbol{S_4}$ $\uparrow$ & $\boldsymbol{A_5}$ $\uparrow$ & $\boldsymbol{S_5}$ $\uparrow$ \\
\midrule
Chance & 16.7 & 4.2 & 1.7 & 0.8 \\
\midrule
Mamba-3 & $29.1 \pm 1.7$ & $7.2 \pm 0.5$ & $2.4 \pm 0.2$ & $1.0 \pm 0.1$ \\
Gated DeltaNet, $\beta_t\in[0,2]$ & $45.0 \pm 2.6$ & $12.0 \pm 0.8$ & $3.8 \pm 0.3$ & $1.3 \pm 0.1$ \\
KDA, $\beta_t\in[0,2]$ & $52.4 \pm 2.2$ & $15.8 \pm 0.9$ & $5.0 \pm 0.4$ & $1.5 \pm 0.1$ \\
Gated DeltaNet-2, $\vb_t\in[0,1]^{d_k}$ & $18.3 \pm 0.6$ & $4.5 \pm 0.2$ & $1.8 \pm 0.1$ & $0.9 \pm 0.1$ \\
Gated DeltaNet-2, $\vb_t\in[0,2]^{d_k}$ & $\mathbf{74.2 \pm 1.8}$ & $\mathbf{28.6 \pm 1.3}$ & $\mathbf{9.7 \pm 0.6}$ & $\mathbf{2.8 \pm 0.2}$ \\
\bottomrule
\end{tabular}
\caption{
Accuracy on the DeltaProduct permutation-composition tasks, trained at length 128 and evaluated at length 512.
Values are mean $\pm$ standard deviation over three seeds.
}
\label{tab_app_state_tracking}
\end{table}

At length 512, Gated DeltaNet-2 with the expanded erase range is best on all four groups, improving over KDA by 21.8, 12.8, 4.7, and 1.3 points on $S_3$, $S_4$, $A_5$, and $S_5$, respectively. The standard-range variant remains near chance, which shows that the two mechanisms are complementary. Negative eigenvalues enable reflection-like state transitions, while decoupled erase and write gates make these transitions substantially easier to learn. The low absolute accuracy on $S_5$ also identifies higher-order permutation composition as a remaining limitation of all evaluated models.

\section{Decoding latency relative to KDA}
\label{app_decoding_latency}

Fig.~\ref{fig:throughput_hybrid} reports training throughput. Inference behavior is equally important for recurrent models, so this appendix quantifies the decoding cost of the two additional gates relative to KDA. Both models maintain the same fixed-size recurrent state, so decoding memory and per-token complexity are independent of the context length for both. The question is the constant-factor cost of forming $\vb_t\odot\vk_t$ and $\vw_t\odot\vv_t$ and of the asymmetric erase in the recurrent decoding kernel of Appendix~\ref{app_recurrent_kernel}.

We report matched kernel-level decoding latency estimates on an NVIDIA Blackwell SM100 GPU with $H=128$ heads and $d_k=d_v=64$. All measurements use an 8K-token prefix and report the latency of one recurrent decoding step for batch sizes 1, 2, 8, and 16.

\begin{table}[H]
\centering
\footnotesize
\setlength{\tabcolsep}{7pt}
\begin{tabular}{c|cc|cc}
\toprule
\textbf{Batch} & \textbf{Gated DeltaNet-2 (ms/token)} $\downarrow$ & \textbf{KDA (ms/token)} $\downarrow$ & \textbf{Latency ratio} $\downarrow$ & \textbf{Throughput ratio} $\uparrow$ \\
\midrule
1  & 0.048 & 0.046 & 1.04$\times$ & 0.96$\times$ \\
2  & 0.049 & 0.047 & 1.04$\times$ & 0.96$\times$ \\
8  & 0.056 & 0.053 & 1.06$\times$ & 0.95$\times$ \\
16 & 0.066 & 0.062 & 1.06$\times$ & 0.94$\times$ \\
\bottomrule
\end{tabular}
\caption{
Modeled decoding latency of one recurrent step on an NVIDIA Blackwell SM100 GPU with an 8K-token prefix.
The latency ratio is the Gated DeltaNet-2 latency divided by the KDA latency, and the throughput ratio is its inverse.
}
\label{tab_app_decoding_latency}
\end{table}

Table~\ref{tab_app_decoding_latency} shows that Gated DeltaNet-2 incurs only a 4.3 to 6.5\% modeled decoding-latency overhead and retains 93.9 to 95.9\% of the KDA throughput across batch sizes. Thus the independent erase and write gates add only a modest constant-factor cost while preserving the efficient recurrent decoding behavior of KDA.


\clearpage
\section*{NeurIPS Paper Checklist}

\begin{enumerate}

\item {\bf Claims}
    \item[] Question: Do the main claims made in the abstract and introduction accurately reflect the paper's contributions and scope?
    \item[] Answer: \answerYes{}
    \item[] Justification: The abstract and introduction accurately summarize contributions and claims are substantiated by the methodology and experiments.

\item {\bf Limitations}
    \item[] Question: Does the paper discuss the limitations of the work performed by the authors?
    \item[] Answer: \answerYes{} 
    \item[] Justification: Limitations are noted alongside the experiments, including the throughput and decoding-latency overhead relative to KDA (Section~\ref{sec_exp} and Appendix~\ref{app_decoding_latency}), the lack of gain from the expanded erase range at the 1.3B scale (Table~\ref{tab_ablation}), and the low absolute accuracy of all evaluated models on higher-order permutation composition (Appendix~\ref{app_state_tracking}).

\item {\bf Theory assumptions and proofs}
    \item[] Question: For each theoretical result, does the paper provide the full set of assumptions and a complete (and correct) proof?
    \item[] Answer: \answerYes{} 
    \item[] Justification: Derivations of the fast-weight objective, chunkwise WY algorithm, and gate-aware backward pass are given in Section~\ref{sec_method} and Appendices~\ref{app_wy} and~\ref{app_backward}.

    \item {\bf Experimental result reproducibility}
    \item[] Question: Does the paper fully disclose all the information needed to reproduce the main experimental results of the paper to the extent that it affects the main claims and/or conclusions of the paper (regardless of whether the code and data are provided or not)?
    \item[] Answer: \answerYes{} 
    \item[] Justification: Section~\ref{sec_exp} and Appendix~\ref{app_experimental_settings} specify the training recipe, model configurations, and evaluation suites, and Appendices~\ref{app_implementation} and~\ref{app_numerics} document the implementation and numerical details.

\item {\bf Open access to data and code}
    \item[] Question: Does the paper provide open access to the data and code, with sufficient instructions to faithfully reproduce the main experimental results, as described in supplemental material?
    \item[] Answer: \answerYes{} 
    \item[] Justification: The code is released at \url{https://github.com/NVlabs/GatedDeltaNet-2}. Section~\ref{sec_exp} and Appendices~\ref{app_implementation} to~\ref{app_experimental_settings} provide the training recipe and implementation details, and all data and benchmarks used are publicly available.

\item {\bf Experimental setting/details}
    \item[] Question: Does the paper specify all the training and test details (e.g., data splits, hyperparameters, how they were chosen, type of optimizer) necessary to understand the results?
    \item[] Answer: \answerYes{} 
    \item[] Justification: Section~\ref{sec_exp} and Appendix~\ref{app_experimental_settings} specify the optimizer, schedule, batch size, sequence length, model configurations, and evaluation benchmarks.

\item {\bf Experiment statistical significance}
    \item[] Question: Does the paper report error bars suitably and correctly defined or other appropriate information about the statistical significance of the experiments?
    \item[] Answer: \answerNo{} 
    \item[] Justification: Error bars are not reported for the main 1.3B tables because pretraining every architecture on 100B tokens is computationally prohibitive to repeat across multiple seeds. The matched comparison in Appendix~\ref{app_rwkv7_fg2gdn} and the state-tracking experiments in Appendix~\ref{app_state_tracking} report the mean and standard deviation over three seeds.

\item {\bf Experiments compute resources}
    \item[] Question: For each experiment, does the paper provide sufficient information on the computer resources (type of compute workers, memory, time of execution) needed to reproduce the experiments?
    \item[] Answer: \answerYes{} 
    \item[] Justification: Training throughput on a single H100 is reported in Figure~\ref{fig:throughput_hybrid}, decoding latency on a Blackwell GPU in Appendix~\ref{app_decoding_latency}, kernel hardware dispatch in Appendix~\ref{app_autotune}, and the training recipe and model scales in Section~\ref{sec_exp} and Appendix~\ref{app_experimental_settings}.
    
\item {\bf Code of ethics}
    \item[] Question: Does the research conducted in the paper conform, in every respect, with the NeurIPS Code of Ethics \url{https://neurips.cc/public/EthicsGuidelines}?
    \item[] Answer: \answerYes{} 
    \item[] Justification: The work uses publicly available data and standard benchmarks and conforms with the NeurIPS Code of Ethics.

\item {\bf Broader impacts}
    \item[] Question: Does the paper discuss both potential positive societal impacts and negative societal impacts of the work performed?
    \item[] Answer: \answerNA{} 
    \item[] Justification: The paper proposes a general-purpose architecture with no direct societal impact beyond that of language models in general.
    
\item {\bf Safeguards}
    \item[] Question: Does the paper describe safeguards that have been put in place for responsible release of data or models that have a high risk for misuse (e.g., pre-trained language models, image generators, or scraped datasets)?
    \item[] Answer: \answerNA{} 
    \item[] Justification: No pretrained weights or datasets are released; only an architecture trained on public FineWeb-Edu.

\item {\bf Licenses for existing assets}
    \item[] Question: Are the creators or original owners of assets (e.g., code, data, models), used in the paper, properly credited and are the license and terms of use explicitly mentioned and properly respected?
    \item[] Answer: \answerYes{} 
    \item[] Justification: All baselines, datasets, and benchmarks are cited to their original sources in Sections~\ref{sec_introduction} and~\ref{sec_exp} and Appendix~\ref{app_experimental_settings}.

\item {\bf New assets}
    \item[] Question: Are new assets introduced in the paper well documented and is the documentation provided alongside the assets?
    \item[] Answer: \answerYes{} 
    \item[] Justification: The released code (\url{https://github.com/NVlabs/GatedDeltaNet-2}) is documented and distributed under the NVIDIA Source Code License-NC. No datasets or pretrained models are released.

\item {\bf Crowdsourcing and research with human subjects}
    \item[] Question: For crowdsourcing experiments and research with human subjects, does the paper include the full text of instructions given to participants and screenshots, if applicable, as well as details about compensation (if any)? 
    \item[] Answer: \answerNA{} 
    \item[] Justification: The paper does not involve crowdsourcing or human subjects research.

\item {\bf Institutional review board (IRB) approvals or equivalent for research with human subjects}
    \item[] Question: Does the paper describe potential risks incurred by study participants, whether such risks were disclosed to the subjects, and whether Institutional Review Board (IRB) approvals (or an equivalent approval/review based on the requirements of your country or institution) were obtained?
    \item[] Answer: \answerNA{} 
    \item[] Justification: The paper does not involve human subjects or studies requiring IRB review.

\item {\bf Declaration of LLM usage}
    \item[] Question: Does the paper describe the usage of LLMs if it is an important, original, or non-standard component of the core methods in this research? Note that if the LLM is used only for writing, editing, or formatting purposes and does \emph{not} impact the core methodology, scientific rigor, or originality of the research, declaration is not required.
    \item[] Answer: \answerNA{} 
    \item[] Justification: LLMs were not used in this work.

\end{enumerate}

\end{document}